\PassOptionsToPackage{table}{xcolor}
\documentclass{article} %
\usepackage[preprint]{neurips_2023}

\usepackage[utf8]{inputenc} %
\usepackage[T1]{fontenc}    %
\usepackage{hyperref}       %
\usepackage{url}            %
\usepackage{amsfonts}       %
\usepackage{nicefrac}       %
\usepackage{microtype}      %
\usepackage{booktabs}

\usepackage{amsmath,amsfonts,bm}

\def\eqref#1{equation~\ref{#1}}

\def\1{\bm{1}}

\DeclareMathAlphabet{\mathsfit}{\encodingdefault}{\sfdefault}{m}{sl}
\SetMathAlphabet{\mathsfit}{bold}{\encodingdefault}{\sfdefault}{bx}{n}

\usepackage{paralist}
\usepackage{wrapfig}

\usepackage{xspace}
\newcommand{\ours}{\textsc{EDA}\xspace}

\usepackage{tabulary}
\usepackage{color, colortbl}
\usepackage[normalem]{ulem}
\usepackage{url}
\usepackage{multirow}
\usepackage{float}
\usepackage{paralist}
\usepackage{comment}
\usepackage{subcaption}

\usepackage[textwidth=0.8in,textsize=tiny]{todonotes}
\usepackage{soul}

\usepackage{makecell}

\newcommand{\user}[1]{\textcolor{gray}{\textit{#1}}}
\newcommand{\error}[1]{\textcolor{red}{\textbf{#1}}}

\usepackage{array}
\newcolumntype{H}{>{\setbox0=\hbox\bgroup}c<{\egroup}@{}}

\definecolor{mycolor}{RGB}{0,0,255}
\newcommand{\scorecolor}[1]{\cellcolor{mycolor!#1}}

\usepackage[capitalize]{cleveref}
\crefformat{equation}{Eq.~(#2#1#3)}
\crefname{section}{§}{§§}
\Crefname{section}{§}{§§}

\usepackage{tcolorbox}

\title{Probing the Multi-turn Planning Capabilities of LLMs via 20 Question Games}

\author{
Yizhe Zhang, \quad Jiarui Lu, \quad Navdeep Jaitly\\
 Apple\\
\texttt{\{yizzhang, jiarui\_lu, njaitly\}@apple.com}
}

\begin{document}
\maketitle

\begin{abstract}
Large language models (LLMs) are effective at answering questions that are clearly asked. However, when faced with ambiguous queries they can act unpredictably and produce incorrect outputs. This underscores the need for the development of intelligent agents capable of asking clarification questions to resolve ambiguities effectively. This capability requires complex understanding, state tracking, reasoning and planning over multiple conversational turns. However, directly measuring this can be challenging.
In this paper, we offer a surrogate problem which assesses an LLMs's capability to deduce an entity unknown to itself, but revealed to a judge, by asking the judge a series of queries. This \textit{entity-deducing game} can serve as an evaluation framework to probe the conversational reasoning and planning capabilities of language models.
We systematically evaluate various LLMs and discover significant differences in their performance on this task. We find that strong LLMs like GPT-4 outperform human players by a large margin. 
We further employ Behavior Cloning (BC) to examine whether a weaker model is capable of imitating a stronger model and generalizing to data or domains, using only the demonstrations from a stronger model. 
We finally propose to use Reinforcement Learning to enhance reasoning and planning capacity of Vicuna models through episodes of game playing, which lead to significant performance improvement. 
We hope that this problem offers insights into how autonomous agents could be trained to behave more intelligently in ambiguous circumstances.
\end{abstract}

\section{Introduction}

In uncertain circumstances, intelligent conversational agents may need to take the initiative to reduce their uncertainty by asking good questions proactively, thereby solving problems more effectively. This requires intricate, interactive, strategic decision-making and reasoning about the agent's next move in a multi-turn conversation. This capability is crucial in various applications, such as multistep task completion, task-oriented chatbots, recommendations, and conversational search.

Traditionally, dialogue systems, including the clarification process, have been achieved by modularizing various aspects of such tasks into sub-tasks such as \textit{natural language understanding}, \textit{state tracking}, \textit{planning (policy learning)}, and \textit{response generation}. 
However, recent advances in LLM-powered systems have made it possible to create an end-to-end pipeline, opening up new possibilities for developing autonomous agents that can complete complex tasks using enhanced planning and memory capabilities. 
Promising works, such as ReAct \citep{yao2022react}, HuggingGPT \citep{shen2023hugginggpt}, AutoGPT \citep{Auto-GPT}, LangChain \citep{langchain-ai}, GPT-Engineer \citep{gpt-engineer} and BabyAGI \citep{babyagi}, have demonstrated significant potential in this field.
These agents require the underlying LLM to retain and recall important information from previous dialogues, resembling the \textit{understanding} and \textit{state tracking} stage. 
They also rely on the LLM to decompose larger tasks into more manageable components, which is analogous to the \textit{planning} stage.
Among them, some approaches (\textit{e.g.}, HuggingGPT) use a \textit{static} planning strategy by first generating the complete plan via LLM and subsequently tackling each subtask. Other approaches (\textit{e.g.}, AutoGPT) adopt a \textit{dynamic} and \textit{interactive} planning strategy, where the generation of each action is conditioned on the outcome of the previous planning steps.

\begin{figure}[ht!]
    \centering
    \includegraphics[width=0.99\linewidth]{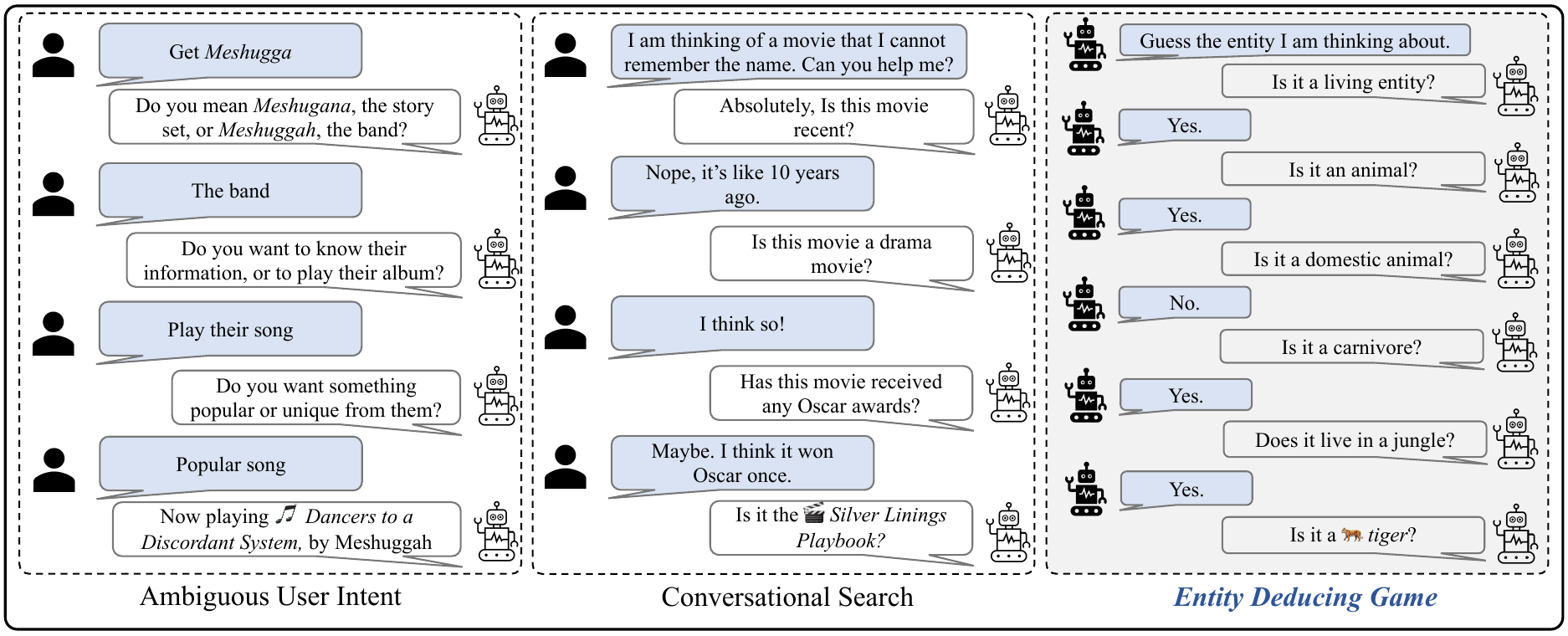}
    \caption{The entity deducing game resembles real scenarios where the agent may need to make strategic decisions regarding the clarification question to be asked based on the current conversation to elicit the actual user intent in as few turns as possible. 
    }
    \label{fig:dialog}
\end{figure}  

While LLM-powered autonomous agents can solve a wide variety of tasks, they can perform poorly when a user's instructions are unclear. This poses a critical challenge -- how to accurately elicit and capture the user's intents, which are often nuanced, unpredictable, and noisy, to enable dynamic rather than static human-computer interactions. For example, in Figure~\ref{fig:dialog},
the agent has three objectives: 
1) accurately assess the current dialog state;
2) eliminate ambiguity in user's intent and satisfy the user demand by asking strategic questions; 
3) ask as few questions as possible.

Progress in this direction is difficult because directly measuring complex understanding, reasoning and planning capabilities is challenging.
In this study, we investigate this somewhat overlooked research problem -- \textit{how good the LLMs are at asking questions and deducing intent.} 
We propose to use entity-deducing games, specifically the 20 questions game (Q20) \citep{akinator}, to assess the complex reasoning and strategic planning capability of LLMs in formulating precise questions/guesses over long conversations (Figure~\ref{fig:dialog}). 
This game requires a model to infer an unidentified entity through a sequence of questions that elicit simple responses of ``Yes'', ``No'' or ``Maybe'' 
with as few queries as possible. To achieve this, the model must be able to track the dialogue state over turns, and use its reasoning and planning skills to effectively partition and narrow down the search scope.

We systematically evaluated several LLMs for their performance on this task, and found significant differences in their conversational reasoning and planning capabilities.
We then investigated a set of research questions to enhance open-source models using demonstrations from high-performing closed-source models.
Finally, drawing inspiration from Reinforcement Learning from Human Feedback (RLHF) \citep{openai2023gpt4}, we show how PPO \citep{schulman2017proximal} can be used to improve vanilla Vicuna \citep{vicuna2023} models using game playing directly. We provide the code and dataset to facilitate future research 
\footnote{\url{https://github.com/apple/ml-entity-deduction-arena}}.
Our findings offer insights into potential future advancements in autonomous agents that proactively lead the conversation.

\section{Related Work}
\paragraph{Evaluation of Complex Reasoning Capabilities of LLMs} There has been extensive work on evaluating the complex reasoning abilities of LLMs\citep{huang2022towards}. Prior work has created diverse benchmarks, like HELM \citep{liang2022holistic} BIG-bench \citep{srivastava2022beyond}, SuperGLUE \citep{sarlin2020superglue}, LAMA \citep{petroni2019language}, and CoT-Hub \citep{fu2023chain}, which have enabled researchers to assess LLMs across a spectrum of tasks involving 
arithmetic and math (GSM8K) \citep{cobbe2021training}, commonsense (ARC) \citep{clark2018think}, symbolic (BBH)~\citep{suzgun2022challenging}, knowledge (MMLU) \citep{hendrycks2020measuring}, coding (HumanEval) \citep{chen2021evaluating}, and factual (SummEdits) \citep{laban2023llms} have been developed to gauge their real-world abilities. 
We aim to introduce a novel benchmark to complement existing benchmarks, by evaluating the model's capacity on multi-turn deductive reasoning, state tracking, fact-recalling and strategical planning.

\paragraph{Evaluation of Planning} Evaluation of the planning abilities of LLMs is relatively rare.
\citet{valmeekam2022large} proposed an assessment framework to gauge the planning capabilities of LLMs to generate valid actions to achieve a special goal and provide a rationale for the plan. 
Their evaluation on 8 planning tasks reveals LLMs, like GPT-3 \citep{brown2020language}, seem to display a dismal performance. \citet{valmeekam2023planning} further evaluates on GPT-4 and suggests the autonomous learning capacity of LLMs to formulate plans is limited and dependent on properly designed heuristics.
\citet{xie2023translating} similarly indicate that LLMs may encounter difficulties in generating planning goals related to numerical or spatial reasoning, while being responsive to the specific prompts used.  
Unlike these studies, our task solely relies on textual representations of goals.

\paragraph{Multiturn benchmarks}
MT-Bench \citep{zheng2023judging} assesses the multi-turn conversation and instruction-following ability of LLMs by annotating their responses to over 80 challenging questions involving writing, role-play, extraction, reasoning, math, coding and knowledge. \citet{bang2023multitask} evaluates LLMs on 23 different datasets encompassing tasks related to logical reasoning, non-textual reasoning, and commonsense reasoning. The study reveals that incorporating interactive features in LLMs can enhance their performance by employing multi-turn prompt engineering. LMRL Gym \citep{abdulhai2023lmrl}, a concurrent work, investigated Multi-Turn Reinforcement Learning with Language Models across multiple benchmark tasks. In contrast, our work focuses on the depth by presenting a thorough analysis of the entity deduction task.

\paragraph{Entity-deduction game} 
Testing the model's ability to deduct an ambiguous entity or asking clarification questions \citep{aliannejadi2019asking, cho2019contrastive} has been utilized as a testbed on dialogue systems and visual reasoning tasks. InfoBot \citep{dhingra2016towards} uses reinforcement learning to learn an optimal dialog policy for identifying movie entries from a movie database.
ESP \citep{von2004labeling} and Peekaboom \citep{von2006peekaboom} demonstrated that deduction games can effectively gather labeled data.
GuessWhat?! \citep{de2017guesswhat} and ReferIt \citep{kazemzade2014referring}, assess the visual reasoning capabilities of tested models.
These benchmarks involve a model seeing an entity-deduction conversation based on an image 
to guess the referred object in the image. 
Our work instead aims to gauge on the model's ability on generating the conversation.

\section{Entity-deduction Arena (\ours)}
During an entity-deducing game session, two players engage in a collaborative game regarding a specific entity. 
One player, the ``\textit{judge}'' (\texttt{J}), is provided with the entity and is expected to respond to queries or guesses from the guesser using only the responses ``Yes,'' ``No,'' or ``Maybe,''.
The other player, referred to as the ``\textit{guesser}'' (\texttt{G}) is agnostic about the entity, and is supposed to pose a series of questions to deduce the entity using as few queries as possible. 

The judge does not require access to the dialogue history and only needs the entity and current question to provide a response \footnote{Our experiment indicate that incorporating entire dialogue history negatively impacts accuracy. The additional information tends to confuse the judge rather than improve understanding.}. This task is akin to closed-book QA \citep{roberts-etal-2020-much}, which state-of-the-art LLMs can reasonably handle.
On the other hand, playing the guesser is more demanding.
A proficient \texttt{G} necessitates several multi-turn dialogue capabilities working in synergy:

\begin{enumerate}
\itemindent=-15pt

   \item \textbf{State Tracking and Understanding}: \texttt{G} must comprehend multi-turn context, track asked questions, and understand its position in the game and coreference resolution.
    
    \item \textbf{Strategic Planning}: \texttt{G} needs to strategically ask questions to progress efficiently towards a better state, avoiding redundant queries and ensuring consistency with prior knowledge.
    
    \item \textbf{Inductive Reasoning}: \texttt{G} must use conversation comprehension to generate conjectures based on acquired knowledge. \texttt{G} must inherently establish a \textit{taxonomy representation} to efficiently and accurately identify the correct entity among numerous options.
\end{enumerate}

Our assessment, referred to as the \textit{Entity-Deduction Arena} (\ours), focuses on evaluating the capability of various LLMs as the guesser, as a proxy to probe their overall capabilities in handling complex multi-turn conversational tasks involving proactively asking clarification questions.

\subsection{Experimental settings}

\paragraph{Datasets} 
We conducted the evaluation on two proposed datasets: \textit{Things} and \textit{Celebrities}. 

The \textit{Things} dataset consists of 500 entities that are commonly found in the web corpus \citep{raffel2020exploring}. It encompasses a wide range of categories, such as common-life objects, animals, foods, plants, vehicles, clothing, professions, materials, instruments, places, birds, sports, buildings, furniture, celestial bodies, mythical creatures, events, activities. The larger category ``objects'' account for 20\% percentage of the total number of items.

The \textit{Celebrities} dataset consists of 500 celebrity names covers 32 different nationality, diverse eras of life and various occupations. Among the nationalities, the United States holds the largest share with 27.7\%. In terms of occupations, the largest categories are actor/actress, musician, and writer, accounting for 18.4\%, 15.8\%, and 14.7\% respectively. 29.6\% of the celebrities belong to the pre-20th century era, while 18.4\% are from the post-21st century era, and the remaining individuals are from the 20th century.
Both datasets were divided into training, evaluation, and testing, with sizes of 300, 100, and 100 respectively. The composition of each dataset is provided in Appendix~\ref{app:dataset}.

\paragraph{Judge (\texttt{J})} We employ GPT-3.5-turbo as the judge. The judge takes the entity, questions from the guesser and the following prompt to generate a response of ``Yes,'' ``No,'' or ``Maybe'' for Things dataset. In guessing the celebrity name, the choices are ``Yes,'' ``No,'' or ``Dunno''. Consequently, the resulting prompt is slightly different. The prompt, and rationale for using ``Dunno'' for the \textit{Celebrities} dataset are provided in Appendix~\ref{app:celeb})

\begin{center}
\begin{tcolorbox}[width=0.99\textwidth]
\textit{Based on your knowledge about the entity: \textcolor{blue}{\{entity\}}, respond to the following question or guess. Limit your respond to only ``Yes.'', ``No.'' or ``Maybe.'', with no explanation or other words. Never say the answer in your response. If the question is to solicit the answer, respond ``No.''. \textcolor{gray}{\textbackslash n}
Question/Guess: \textcolor{blue}{\{question\}} (Yes/No/Maybe)
}
\end{tcolorbox}
\end{center}

Whenever the correct answer is contained in the generation from \texttt{G} as an exact substring match, we manually set the output of \texttt{J} to be \textit{``Bingo!''} and \texttt{G} wins this game. 
At the penultimate step of \texttt{J}, an extra prompt \textit{``You must guess now, what's it?''} will be appended to \texttt{J}'s response to guide \texttt{G} in making the ultimate guess. 
To emulate more deterministic responses from \texttt{J}, we use a temperature of 0.2 for the generation.
Admittedly, we observe that the judge model exhibits occasional inaccuracies in its responses, resulting in a noisy environment. We asked internal non-paid volunteers to annotate 300 responses from \texttt{J} for each dataset, and the error rate is 3.07\% and 2.95\% for \textit{Things} and \textit{Celebrities}, respectively. 
Nevertheless, this noisy environment of \texttt{J} is consistent with all models.

\paragraph{Guesser (\texttt{G})} 
The guesser model is unaware of the entity. It receives the current dialogue history as input and generates the next question or final guess, guided by the instructions provided in the following (the prompt for Celebrities is provide in Appendix~\ref{app:celeb}):
\begin{center}
\begin{tcolorbox}[width=0.99\textwidth]
\textit{Your task is to ask a series of questions to deduce the entity that I'm thinking of with as few queries as possible. Only ask questions that can be answered by ``Yes,'' ``No,'' or ``Maybe''. Do not ask for hint. Make your question brief with no linebreaker. Now start asking a question.
\textcolor{gray}{\textbackslash n}
\textcolor{blue}{\{dialog history\}}
}
\end{tcolorbox}
\end{center}
We used a sampling approach with a temperature of 0.8 for all experiments, which sufficiently yields diverse outcome. The conversation format for each LLMs is based on their official guidelines.

\paragraph{Evaluation metrics} We assess the model's performance by evaluating its final prediction using the Exact Match (EM) criteria \footnote{Alternatively, a more lenient evaluation metric could be employed, \textit{e.g.} embedding similarities or LLM judges. Nevertheless, we discovered that these metrics rely on the specific embedding model or LLM judge employed, potentially resulting in less consistent and reliable evaluations than the EM criteria.}. This evaluation considers four key metrics: 
1) \textbf{\#Turns}, which represents the average number of turns taken to complete the game. Games terminate at 20 turns if failed. Lower is better. 
2) \textbf{Success} rate, which indicates the percentage of games won by \texttt{G}. Higher is better.
3) \textbf{\#Yes}, which represents the average number of ``yes'' responses received from the \texttt{J}.
4) \textbf{Score}, which is a combined game score of \textbf{\#Turns} and success rate, defined in~\cref{eq:score}. 
\begin{equation}
S =
\begin{cases}
1 - \lambda \cdot \max(\textbf{\#Turns} - 5, 0) & \text{if \texttt{G} wins,} \\
0 & \text{if \texttt{G} loses,}
\end{cases}
\label{eq:score}
\end{equation}

where 
$\lambda=0.02$ is a balancing hyperparameter. A higher value for $\lambda$ would result in assigning inadequate rewards for successful games that take longer, while a smaller value would penalize the model too little for taking a longer turns. In our experiments, we discovered that a value of 0.02 led to models with a reasonable and stable reward. It is worth mentioning that reward design is more of a heuristic, and there may be superior choices available. 
The \textbf{\#Yes} is more of a statistic than a evaluation metric, but we have observed some correlation between this metric and the final performance. Intuitively, a losing game is often characterized by a high frequency of unproductive guesses (with ``No'' or ``Maybe'' response from \texttt{J}).

\section{Benchmarking LLMs on \ours}

\begin{table}[ht!]
\scriptsize
\centering
\rowcolors{2}{gray!25}{white} %
\begin{tabular}{lcccccccc}
    \toprule
    & \multicolumn{4}{c}{\textbf{Things}} & \multicolumn{4}{c}{\textbf{Celebrities}} \\
    \cmidrule(r){2-5} \cmidrule(r){6-9}
     & \textbf{\#Turns ($\downarrow$)} & \textbf{Success  ($\uparrow$)} & \textbf{\#Yes} & \textbf{Score ($\uparrow$)} & \textbf{\#Turns ($\downarrow$)} & \textbf{Success ($\uparrow$)} & \textbf{\#Yes} & \textbf{Score ($\uparrow$)} \\
    \midrule
    GPT-4 & \textbf{17.8$\pm$0.2} & \textbf{0.31$\pm$0.03} & 5.9$\pm$0.1 & \scorecolor{13}\textbf{0.26$\pm$0.02} &  \textbf{17.3$\pm$0.1} & \textbf{0.50$\pm$0.02} & 6.8$\pm$0.2 & \scorecolor{20}\textbf{0.40$\pm$0.02} \\
    GPT-3.5 & 18.2$\pm$0.2 & 0.28$\pm$0.02 & 6.3$\pm$0.1 & \scorecolor{11.5}0.23$\pm$0.02 & 18.8$\pm$0.2 & 0.27$\pm$0.03 & 7.4$\pm$0.2 & \scorecolor{10.5}0.21$\pm$0.03 \\
    Claude-2 & 18.4$\pm$0.3 & 0.21$\pm$0.03 & 5.0$\pm$0.1 & \scorecolor{9}0.18$\pm$0.03 & 17.6$\pm$0.2 & 0.31$\pm$0.02 & 5.6$\pm$0.1 & \scorecolor{13}0.26$\pm$0.02 \\
    Claude-1 & 18.8$\pm$0.1 & 0.16$\pm$0.02 & 4.2$\pm$0.1 & \scorecolor{6.5}0.13$\pm$0.02 & 17.7$\pm$0.2 & 0.29$\pm$0.03 & 5.3$\pm$0.2 & \scorecolor{12.5}0.25$\pm$0.02 \\
    Vicuna 13B & 18.4$\pm$0.1 & 0.18$\pm$0.02 & 5.0$\pm$0.2 &\scorecolor{7.5}0.15$\pm$0.02 & 18.7$\pm$0.2 & 0.22$\pm$0.03 & 6.1$\pm$0.1 &\scorecolor{9}0.18$\pm$0.02 \\
    Vicuna 7B & 19.5$\pm$0.2 & 0.09$\pm$0.02 & 5.7$\pm$0.2 & \scorecolor{3.5}0.07$\pm$0.02 & 19.6$\pm$0.3 & 0.06$\pm$0.02 & 5.9$\pm$0.2 & \scorecolor{2}0.05$\pm$0.02 \\
    Mistral 7B & 18.9$\pm$0.1 & 0.13$\pm$0.02 & 3.8$\pm$0.5 & \scorecolor{5.5}0.11$\pm$0.02 & 18.2$\pm$0.1 & 0.22$\pm$0.04 & 4.3$\pm$0.1 & \scorecolor{10}0.20$\pm$0.03 \\
    \midrule
    V-FT 7B (All) & 19.2$\pm$0.1 & 0.13$\pm$0.01 & 6.1$\pm$0.1 & \scorecolor{5}0.10$\pm$0.01 & 19.3$\pm$0.1 & 0.16$\pm$0.02 & 7.6$\pm$0.3 & \scorecolor{6.5}0.13$\pm$0.02 \\
    V-FT 7B (Suc.) & 18.0$\pm$0.1 & 0.23$\pm$0.01 & 5.1$\pm$0.2 & \scorecolor{10}0.20$\pm$0.01 & 19.0$\pm$0.2 & 0.15$\pm$0.02 & 6.3$\pm$0.2 & \scorecolor{6.5}0.13$\pm$0.02 \\
    V-FT 13B (All) & 18.6$\pm$0.2 & 0.21$\pm$0.03 & 6.1$\pm$0.2 & \scorecolor{8.5}0.17$\pm$0.02  & 18.8$\pm$0.2 & 0.22$\pm$0.01 & 6.2$\pm$0.2 & \scorecolor{9}0.18$\pm$0.01 \\
    V-FT 13B (Suc.) & 18.0$\pm$0.2 & 0.25$\pm$0.02 &  4.5$\pm$0.1 & \scorecolor{10.5}0.21$\pm$0.03  & 18.4$\pm$0.3 & 0.23$\pm$0.04 & 5.9$\pm$0.2 & \scorecolor{9.5}0.19$\pm$0.03 \\
    \midrule
    V-RLGP 7B &17.8$\pm$0.1 & 0.26$\pm$0.02 & 4.7$\pm$0.1 & \scorecolor{11}0.22$\pm$0.01&		18.8$\pm$0.1 &	0.16$\pm$0.01&	5.9$\pm$0.1&	\scorecolor{7}0.14$\pm$0.00 \\
    V-RLGP 13B & 17.9$\pm$0.1 & 0.27$\pm$0.02 & 4.5$\pm$0.1 &\scorecolor{11.5}0.23$\pm$0.01&	18.5$\pm$0.2	&0.26$\pm$0.03	&6.1$\pm$0.1	&\scorecolor{10.5}0.21$\pm$0.02 \\
    \bottomrule
\end{tabular}
\vspace{4mm}
\caption{Benchmark of LLMs on the \ours datasets (\textit{Things}, \textit{Celebrities}). \textbf{\#Turns} denotes the average number of turns taken. \textbf{Success} denotes the percentage of the games that the \texttt{G} wins. \textbf{\#Yes} denotes the average number of ``yes'' response received from \texttt{J}. \textbf{Score} indicates the reward score defined in ~\cref{eq:score}. 
The mean and standard deviation were calculated based on 5 repetitions.
}
\label{tab:llm_eval}
\end{table}

We assess several widely-used LLMs, such as GPT-4, GPT-3.5, Claude-1/2, Vicuna (7B-v1.3, 13B-v1.3)~\citep{vicuna2023} and Mistral-7B~\citep{jiang2023mistral} using our \ours benchmarks. Unfortunately, we are unable to evaluate Llama-2 or its derivatives due to license restrictions.
Llama-1~\citep{touvron2023llama} is also excluded from the comparison as it lacks a conversation finetuned model, and as such it often does not adhere to our instructions.
The corresponding results are presented in Table~\ref{tab:llm_eval}. 
The means and standard deviations for each dataset and LLM were calculated using 30 items from 5 separate runs.

In general, GPT-4 attained the best performance among the models evaluated. Around 1/3 of the entities were accurately predicted.
The runner-ups are ChatGPT-3.5 and Claude-2. ChatGPT-3.5 is superior on the \textit{Things} datasets, while Claude-2 shows better strength on the \textit{Celebrities} datasets.
Despite being much smaller, open-source models such as the Vicuna 13B and Mistral 7B model performed comparably well, not trailing too far behind the closed-source models. In particular, Vicuna 13B outperformed Claude-1 when it comes to \textit{Things}. This suggests that open-sourced models have significant potential. Naturally, models with a higher success rate usually require fewer turns.

It is plausible that a small portion of entity deduction game sessions may exist on the web and were a part of the pretraining data. However, these sessions likely contribute only a minuscule fraction.
It is unlikely that all the entities utilized in our study have appeared in the training data. Nevertheless, robust models such as GPT-4 exhibit consistently strong performance across numerous tested entities. 

We have also conducted a large-scale human baseline in a subset of 30 tested examples for each dataset. The discussion of these results can be found in Appendix~\ref{app:human_baseline}.

\begin{table}[ht!]
    \centering
    \scriptsize
    \begin{tabular}{p{2.3in}|p{2.3in}}
    \toprule
    \thead{\small Successful game} & \thead{\small Planning issue (Early Enumeration)} \\
    \hline
    \makecell[l]{\texttt{G}: Is it a living entity?  \user{\texttt{J}: No.}\\ \texttt{G}: Is it man-made?  \user{\texttt{J}: Yes.}\\ \texttt{G}: Can it be held in a single hand?  \user{\texttt{J}: Yes.}\\ \texttt{G}: Is it electronic?  \user{\texttt{J}: Yes.}\\ \texttt{G}: Is it used for communication?  \user{\texttt{J}: Maybe.}\\ \texttt{G}: Can it store information?  \user{\texttt{J}: Yes.}\\ \texttt{G}: Is it a portable device?  \user{\texttt{J}: Maybe.}\\ \texttt{G}: Is it commonly found in homes?  \user{\texttt{J}: Yes.}\\ \texttt{G}: Is it related to entertainment?  \user{\texttt{J}: No.}\\ \texttt{G}: Is it used for work or productivity?  \user{\texttt{J}: Yes.}\\ \texttt{G}: Is it a type of computer or computing device?  \user{\texttt{J}: Maybe.}\\ \texttt{G}: Is it used for creating or editing documents?  \user{\texttt{J}: Yes.}\\ \texttt{G}: Is it a keyboard?  \user{\texttt{J}: No.}\\ \texttt{G}: Is it a mouse?  \user{\texttt{J}: No.}\\ \texttt{G}: Is it a printer?  \user{\texttt{J}: \textcolor{blue}{Bingo!}}}&
    \makecell[l]{\texttt{G}: Is the entity a living thing?  \user{\texttt{J}: No.}\\ \texttt{G}: Is the entity an object?  \user{\texttt{J}: Yes.}\\ \texttt{G}: Is the object a tool or a device?  \user{\texttt{J}: Yes.}\\ \texttt{G}: Is the tool or device used for \error{communication}?  \user{\texttt{J}: No.}\\ \texttt{G}: Is the tool or device used for \error{transportation}?  \user{\texttt{J}: No.}\\ \texttt{G}: Is the tool or device used for \error{construction}?  \user{\texttt{J}: No.}\\ \texttt{G}: Is the tool or device used for \error{entertainment}?  \user{\texttt{J}: No.}\\ \texttt{G}: Is the tool or device used for \error{cooking}?  \user{\texttt{J}: No.}\\ \texttt{G}: Is the tool or device used for \error{cleaning}?  \user{\texttt{J}: No.}\\ \texttt{G}: Is the tool or device used for \error{protection}?  \user{\texttt{J}: No.}\\ \texttt{G}: Is the tool or device used for \error{sports or athletics}?  \user{\texttt{J}: No.} \\ ...} \\
    \toprule
    \thead{\small Planning issue (Redundancy)} & \thead{\small Reasoning issue (Inconsistency)} \\
    \hline
    \makecell[l]{\texttt{G}: Is the entity a person?  \user{\texttt{J}: No.}\\ \texttt{G}: Is the entity a living thing?  \user{\texttt{J}: No.}\\ \texttt{G}: Is the entity a thing?  \user{\texttt{J}: Yes.}\\ \texttt{G}: Is the entity an object?  \user{\texttt{J}: Yes.}\\ \texttt{G}: Is the entity a material object?  \user{\texttt{J}: Yes.}\\ \texttt{G}: Is the entity a solid object?  \user{\texttt{J}: Yes.}\\ \texttt{G}: Is the entity a natural object?  \user{\texttt{J}: No.}\\ \texttt{G}: Is the entity an \error{artificial object}?  \user{\texttt{J}: Yes.}\\ \texttt{G}: Is the entity a \error{man-made object}?  \user{\texttt{J}: Yes.}\\ \texttt{G}: Is the entity an \error{inanimate object}?  \user{\texttt{J}: Yes.}\\ \texttt{G}: Is the entity a machine?  \user{\texttt{J}: Yes.} \\ ...}&
    \makecell[l]{\texttt{G}: Is the entity a living being?  \user{\texttt{J}: No.}\\ \texttt{G}: Is the entity an object or a thing?  \user{\texttt{J}: Yes.}\\ \texttt{G}: Is the object typically found indoors?  \user{\texttt{J}: Yes.}\\ \texttt{G}: Is the object used primarily for a specific purpose?  \user{\texttt{J}: Yes.}\\ \texttt{G}: Is the object small and portable?  \user{\texttt{J}: Maybe.}\\ \texttt{G}: Is the object electronic?  \user{\texttt{J}: Yes.}\\ \texttt{G}: Is the object a device used for communication  \user{\texttt{J}: No.}\\ \texttt{G}: Is the object a device used for \error{entertainment} or media  \user{\texttt{J}: No.}\\ \texttt{G}: Is the object a device used for productivity or work  \user{\texttt{J}: Yes.}\\ \texttt{G}: Is the object a laptop?  \user{\texttt{J}: No.}\\ \texttt{G}: Is the object a \error{game console}?  \user{\texttt{J}: No.} \\ ...} \\
    \bottomrule
    \end{tabular}
    \vspace{2mm}
    \caption{The gameplay with the hidden entity ``printer''. The successful and problematic gameplay scenarios are generated using GPT-4 and vicuna-7B model as \texttt{G}, respectively.}
    \label{tab:failure_mode}
\end{table}

We present several real game runs in Table~\ref{tab:failure_mode} on \textit{Things}. Examples on \textit{Celebrities} are in Appendix~\ref{app:failure_mode}. We observe that an effective game strategy typically featured by narrowing down the scope through initial questions that divide the remaining search space into two halves of similar capacity, similar to a \textit{binary tree search} process. During the late game when it becomes challenging to find good binary-partitioning questions, the model start enumerates a smaller number of possibilities.
Interestingly, the more advanced models like GPT-4 seem to possess this capability to a decent extent, while weaker models exhibit different failure modes in planning and reasoning abilities (Table~\ref{tab:failure_mode}): 1) (\textbf{Early Enumeration}) initiating enumeration too early without first posing high-level questions to narrow down the scope; 2) (\textbf{Redundancy}) asking questions similar to previous ones; 3) (\textbf{Inconsistency}) asking questions or making guesses that are inconsistent to the answers received in previous turns.

We also compare several games between GPT-4 and human players in Appendix~\ref{app:GPT}, highlighting that strong LLMs share similarities with human players in terms of their ability to \textit{backtrack} when they are on the wrong path. One advantage of LLM over humans is that LLM may have a superior level of knowledge (Appendix~\ref{app:GPT} Table~\ref{tab:gptvshuman_celeb}), empowering them to ask questions using \textit{tacit knowledge} among LLMs.
One weakness of the LLMs comparing to humans is that LLMs tend to enumerate or repeat to follow the same pattern of previous turns, demonstrating \textit{self-reinforced} undesirable behaviors. 
Weaker models tend to be attracted to this absorbing state of repetition or artifacts, and accumulate errors until they ultimately fail (Appendix~\ref{app:failure_mode}). 
Stronger models occasionally display the ability to escape from this repetitive patterns and rectify early errors, which presumably be attributed to their stronger adherence to the high-level plan.

We then address the following Research Questions (RQs): 

\begin{table}[ht!]
    \centering
    \scriptsize
    \begin{tabular}{p{0.1 in}|p{3.2in}|p{1.8in}}
    \toprule
    \thead{\#} & \thead{Current top 5 guesses before generating question} & \thead{Generated question}\\
    \hline
    \makecell[l]{
    1 \\
    2 \\
    3 \\
    4 \\
    5 \\
    6 \\
    7 \\
    8 \\
    9 \\   
    10 \\
    11 \\
    12 \\
    13 \\   
    14 \\
    \\
    15 \\
    } &
    \begin{subtable}{0.9\linewidth}
    \begin{tabular}{c|c|c|c|c}
    computer & cellphone & book & car & house\\
    book & computer & car & building & smartphone \\
    computer & {car} & {building} & book & smartphone \\
    smartphone & book & pen & key & mug \\
    {book} & {pen} & key & mug & deck of cards \\
    {key} & deck of cards & {coin} & paper clip & dice \\
    paper book & {wooden spoon} & plastic toy & glass cup & {porcelain plate} \\
    book & playing card & {toothbrush} & {comb} & {paper clip} \\
    deck of cards & book & puzzle & board game & Rubik's cube \\
    book & deck of cards & dice & puzzle &{board game}  \\
    deck of cards& {book} & {paintbrush} & {puzzle} & {dice} \\
    Harmonica & {Guitar pick} & Ocarina & Castanets & Recorder \\
    {Harmonica} & {Recorder} & {Kazoo} & {Maracas} & {Tambourine} \\
    Ukulele & {Violin} & \textcolor{blue}{Guitar} & Mandolin & {Harp} \\
    &&&&\\
    \textcolor{blue}{Guitar} & {Ukulele} & {Banjo} & {Mandolin} & {Lute} \\
    \end{tabular}
    \end{subtable} &
    \makecell[l]{\texttt{G}: Is it a living entity?  \user{\texttt{J}: No.}\\ \texttt{G}: Is it man-made?  \user{\texttt{J}: Yes.}\\ \texttt{G}: Can it be held in one hand? \user{\texttt{J}: Yes.}\\ \texttt{G}: Is it electronic?  \user{\texttt{J}: No.}\\ \texttt{G}: Is it used for writing or drawing?  \user{\texttt{J}: No.}\\ \texttt{G}: Is it made of metal?  \user{\texttt{J}: No.}\\ \texttt{G}: Is it used in cooking or eating? \user{\texttt{J}: No.}\\ \texttt{G}: Is it used for entertainment or leisure?  \user{\texttt{J}: Yes.}\\ \texttt{G}: Is it associated with a specific sport?  \user{\texttt{J}: No.}\\ \texttt{G}: Is it used in board games?  \user{\texttt{J}: No.}\\ \texttt{G}: \error{Is it associated with music?}  \user{\texttt{J}: Yes.}\\ \texttt{G}: Is it a musical instrument?  \user{\texttt{J}: Yes.}\\ \texttt{G}: \error{Is it a stringed instrument?}  \user{\texttt{J}: Yes.}\\ \texttt{G}: Is it commonly used in a symphony \\ \quad  orchestra?  \user{\texttt{J}: No.}\\ \texttt{G}: Is it a guitar?  \user{\texttt{J}: Bingo!}}
    \\
    \bottomrule
    \end{tabular}
    \vspace{2mm}
    \caption{Probing the top 5 ranked guesses made by GPT-4 during a game of guessing ``Guitar''. 
    Some guesses were truncated for readability. 
    \error{Highlighted turns} are the key turns that eliminates all top 5 guesses. 
    }
    \vspace{-3mm}
    \label{tab:dialog-state}
\end{table}

\paragraph{RQ1: How does the LLM generate strategies?}
We conducted an analysis to probe the internal dialog states underneath the LLMs, to understand how these states evolve throughout the game, and how do they affect the strategy of the model. 
To this end, at each turn before the model asks a question, we prompt the LLM to predict top 5 candidates given the current dialogue history.
This probing process is separate from 
(thus will not affect) the gameplay. The probing prompt is provided in Appendix~\ref{app:dialog-state}. A GPT-4 game session of guessing ``guitar'' with dialog state is shown in Table~\ref{tab:dialog-state}. A session on \textit{Celebrities} dataset of guessing ``LeBron James'' is shown in Table~\ref{tab:gpt_celeb_top_k} in Appendix~\ref{app:dialog-state}.
 
From Table~\ref{tab:dialog-state}, it is evident that the model seeks a strategy to ask questions to effectively partition the space among the most promising candidates in the initial 10 turns. After the question is answered, the ranked list of top guesses is adjusted, with certain candidates being eliminated and new candidates being added. For instance, 
during the third turn, the model asks about the item's feasibility of being held in one hand, enabling the elimination of the larger objects in the leaderboard upon receiving an affirmative answer. 
during the fourth turn, the model asks if the item is electronic, which ultimately eliminates the top-ranked guess, ``smartphone'', after receiving a negative response. The remaining items are shifted one position forward to include a new item ``deck of cards'' to the top-5 guesses. 
 
   Interestingly, it is not necessary for the model to always ask questions in order to optimally partition the top-5 candidates. We observed that the strategy also depends on factors including the ease with which a question can be asked to partition the space, and the level of uncertainty the model has about the current top predictions. We provide details on the measurement of uncertainty in Appendix~\ref{app:uncertain}. In situations where the model is uncertain, it may occasionally \textit{backtrack} and \textit{reexamine} categories that were previously missed. For instance, during the 11th turn, question was asked that ruled out all of the top candidates. This could be due to the fact that the top items are similar in nature, and the model realize there is a significant proportion of other classes that have been overlooked. The successful questioning in these cases led to the recovery of these overlooked classes.

\begin{wraptable}{r}{0.55\textwidth}
\centering
\begin{tabular}{lcccc}
    \toprule
    & \textbf{Things ($\uparrow$)} & \textbf{Celebs ($\uparrow$)} \\
    \midrule
    GPT-4 $\rightarrow$ Vicuna 7B & \scorecolor{5}0.10$\pm$0.02 & \scorecolor{7}0.14$\pm$0.04 \\
    GPT-4 $\rightarrow$ GPT-4 & \scorecolor{13}\textbf{0.26$\pm$0.02} & \scorecolor{20}\textbf{0.40$\pm$0.02} \\
    Vicuna 7B $\rightarrow$ Vicuna 7B & \scorecolor{3.5}0.07$\pm$0.02 & \scorecolor{2.5}0.05$\pm$0.02 \\
    Vicuna 7B $\rightarrow$ GPT-4 & \scorecolor{4}0.08$\pm$0.01 & \scorecolor{3}0.06$\pm$0.01 \\
    \bottomrule
\end{tabular}
\caption{Ablation on planning and reasoning ability. Numbers are the game scores with 5 repetitions. X $\rightarrow$ Y indicates using the conversation trajectory from X and use the Y for the last turn. 
}
\label{tab:planning_reasoning}
\end{wraptable}

To summarize, a strong \texttt{G} typically exhibits the following behavior: 1) \textbf{prioritizing high-level questions} before addressing specific details and enumerations, 2) being aware of the current state and asking questions to effectively \textbf{bi-partition the search space}, and 3) being able to \textbf{occasionally backtracking} to consider previously overlooked options.

Additionally, we noticed a high level of consistency in GPT-4's gameplay strategy across multiple repetitions (see example in Appendix~\ref{app:GPT_fail}), despite some fluctuation in the order of the questions asked. This suggests that GPT-4 may rely on its own implicit \textit{taxonomy representation} of entities to make decisions, which remains consistent throughout.

\paragraph{RQ2: Which one is more important in this task, planning or reasoning?}
Planning and reasoning abilities affects different stages of game play. Early questions require careful planning to efficiently partition the space, while late game requires more deductive reasoning skills to make an educated guess.
We consequently assume that the last turn would only require reasoning ability as no strategic move is needed.  
With this assumption, we designed the following experiment to investigate the model's planning and reasoning ability in a finer granularity.
Given a stronger model GPT-4 and a weaker model Vicuna 7B, and their respective game play trajectories, we only replay the last turn in each trajectory by swapping the guesser model \texttt{G}.

The results are presented in Table~\ref{tab:planning_reasoning}.
Comparing the GPT-4 $\rightarrow$ Vicuna 7B with Vicuna 7B, we observe that stronger planning ability from GPT-4 results in significant improvement. 
Regarding different reasoners, Vicuna 7B $\rightarrow$ GPT-4 does not show much improvement over Vicuna 7B. This indicates that planning deficiency could result in an unproductive trajectory, poses significant challenges for reasoning during the final step. Moreover, GPT-4 $\rightarrow$ Vicuna 7B shows regression over GPT-4, emphasizing the importance of reasoning in addition to a strong planning capability. Therefore, it is crucial for both planning and reasoning abilities to be strong and work in synergy to achieve optimal performance. See the full table including the results using human trajectory in the Appendix~\ref{app:reason_vs_plan}.

\section{Enhancing the Open-source Models via Behavior Cloning} 
We further investigate the potential of open-source models.
We used Behavior Cloning (BC) to distill capabilities exhibited by the stronger models into the smaller Vicuna models.
We first collected game demonstrations from the GPT-3.5 model over all 600 training examples from \textit{Things} and \textit{Celebrities}.
We then fine-tuned the Vicuna models using these demonstrations. The experimental settings are provided in the Appendix~\ref{app:FT}. 

\paragraph{RQ3: Can smaller open-source models benefit from imitating larger close-source models?} 
We first fine-tuned the Vicuna 7B model using a total of 600 training instances from both datasets. The results are presented in Table~\ref{tab:llm_eval} as the \textbf{V-FT 7B (All)} model. This yielded a more than 70\% improvement in both datasets, indicating that weaker models can follow the stronger models' demonstration to use a more effective strategy to steer the conversation on unseen entities or names. We have also observed that BC training can effectively mitigate undesirable behavior such as repetition.

\paragraph{RQ4: Should models learn from successful demonstration exclusively?}
To answer this, we selected all the game runs of the entities or names where the teacher model finally won the game from both datasets. This subset was then used to train the Vicuna 7B model, similar to the \textit{Rejection Sampling} \citep{touvron2023llama2} . The results, referred to as \textbf{V-FT 7B (Suc.)} in Table~\ref{tab:llm_eval}, show that imitating from successful experiences outperforms imitating from all experiences (\textbf{V-FT 7B (All)}). It gets close to the teacher model (GPT-3.5) on \textit{Things}. However, when it comes to  the \textit{Celebrities} dataset, the improvement is more limited.

\paragraph{RQ5: Does the model size matter?}
We further compared the performance improvement achieved through finetuning for both Vicuna 7B and 13B models.
The model finetuned on Vicuna 13B, \textbf{V-FT 13B (Suc.)}, also achieved much higher score than Vicuna 13B.
However, the improvement was less substantial than the improvement on \textbf{V-FT 7B (Suc.)}. Again, the improvement over the \textit{Celebrities} dataset is marginal. We hypothesize that guessing the celebrity requires a more case-specific strategies compared to guessing things, thus likely to be more challenging to be superficially imitated.

\section{Reinforcement Learning from Game-play (RLGP)}
We conducted further experiments to investigate whether the performance could be further enhanced through learning solely from the model's own experience.
Drawing inspiration from Reinforcement Learning from Human Feedback (RLHF), we employed Proximal Policy Optimization (PPO) \citep{schulman2017proximal} to train the model by playing with with the judge, \texttt{J}, a technique we refer to as RLGP. We made modifications to the trlX repository \citep{trlx-library}
to facilitate RLGP training.
During training, we assigned the reward defined in~\cref{eq:score}, to the final turn of each rollout. Additionally, we assigned an intermediate reward to turns that received a ``Yes'' response from \texttt{J}. We trained the V-FT 7B (Suc.) and V-FT 13B (Suc.) models in Table~\ref{tab:llm_eval} on 300 \textit{Things} training dateset. Further experimental details can be found in Appendix~\ref{app:RLGP}.
The results in Table~\ref{tab:llm_eval} present the performance of the RL-trained models, denoted as \textbf{V-RLGP}.
\paragraph{V-RLGP} models exhibit improvement compared to the vanilla Vicuna models when tested on the in-domain dataset \textit{Things}. \textbf{V-RLGP 13B} seems to unlock the potential of the Vicuna model, outperforming the \textbf{V-FT 13B} model, matching the performance of the runner-up GPT-3.5. \textbf{V-RLGP 7B} also show substantial improvement over \textbf{V-FT 7B} model.
On the out-domain dataset \textit{Celebrities}, interestingly, \textbf{V-RLGP} models achieves some improvement.

\section{Breakdown Analysis: Do the models agree on successful predictions?}
\begin{figure}[ht!]
    \centering
    \includegraphics[width=1.0\linewidth]{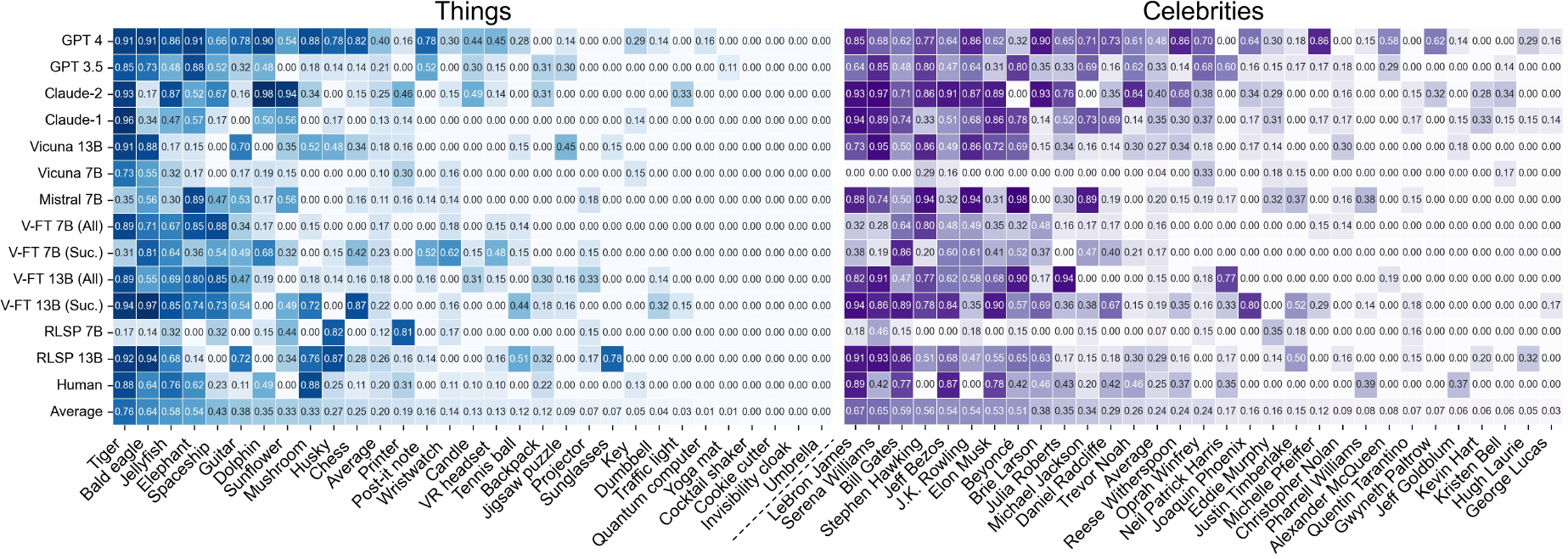}
    \caption{A breakdown of the score of each model on the evaluated items, with the x-axis representing the order of difficulty ranging from easy to difficult. Scores are averaged over 5 repetitions. 
    }
    \label{fig:breakdown}
\end{figure} 

We present a comprehensive breakdown analysis of each model's performance on a subset of 30 evaluated items for each dataset in Figure~\ref{fig:breakdown}.
Within the evaluated items of \textit{Things}, four items 
consistently defy successful predictions by all models.
On the other hand, \textit{Celebrities} exhibits a more scattered pattern, with each celebrity being correctly predicted by at least one model. 

There are correlations between the entities or names that each model can correctly identify, but different models exhibit their own strengths on different subsets. 
For instance, the \textbf{RLGP 13B} model can accurately predict ``Sunglasses'', whereas the stronger models like GPT-4 consistently fails. We also provide some case studies including why GPT-4 consistently fails on ``Yoga mat''
in Appendix~\ref{app:GPT_fail}.

We observed that RLGP models tend to strengthen the performance on items that vanilla models occasionally succeed in, thereby improving their success rate on these specific items. However, RLGP models do not effectively facilitate learning about new items. Conversely, BC fine-tuning appears to excel in achieving success on new items.
Interestingly, the BC fine-tuned models displays different strengths compared to both the initial checkpoint and the expert whose demonstration it mimics. For example, the \textbf{V-FT 13B (Suc.)} model achieves high accuracy in identifying Post-it Note and VR headset, whereas neither the Vicuna 13B nor the GPT-3.5 performs as well in this regard.
\section{Conclusion}

Motivated by a need to develop agents capable of effectively addressing ambiguous user intents, we introduce a testbed for evaluating LLM's strategic planning and deductive reasoning abilities in asking entity-deducing questions. Our findings indicate that SOTA LLMs are able to maintain an intrinsic taxonomic representation of knowledge entities to a certain extent. We further show that this capability can be enhanced through Behavior Cloning or Reinforcement Learning, revealing great potential for further advancements. In future research, we intend to investigate whether the implementation of CoT prompting can further enhance the model's performance in related tasks.

\section*{Acknowledgement}
We thank Yinfei Yang, Xin Zhan, Aonan Zhang, Jiatao Gu, Shuangfei Zhai, Zhe Gan, Bogdan Mazoure, Ruoming Pang, Guoli Yin, Edouard Grave and David Grangier for their valuable feedback and constructive input on our research. Additionally, we extend our thanks to
Abhi Rawat,
Ash Chronister,
Barry Theobald,
Christopher Klein,
David Harrison,
David Koski,
Denise Hui,
Donghao Ren,
Federico Danieli,
Haiming Gang,
Iman Mirzadeh,
Keen You,
Marc Kodama,
Marco Dinacci,
Rangarajan Padur Sivaraman,
Richard Bai,
Sungwook Moon,
Stefan Ludwig,
Vimal Thilak,
Yusu Qian
and all the internal volunteer testers for their exceptional contribution to our interactive human evaluation.
\newpage

{\small
    \bibliography{q20_arxiv}
    \bibliographystyle{iclr2024_conference}
}

\appendix

\onecolumn
\begin{center}
    {\Large \bf Appendix}
\end{center}

\section{Dataset composition}
\label{app:dataset}
\begin{figure}[ht!]
    \centering
    \includegraphics[width=0.9\linewidth]{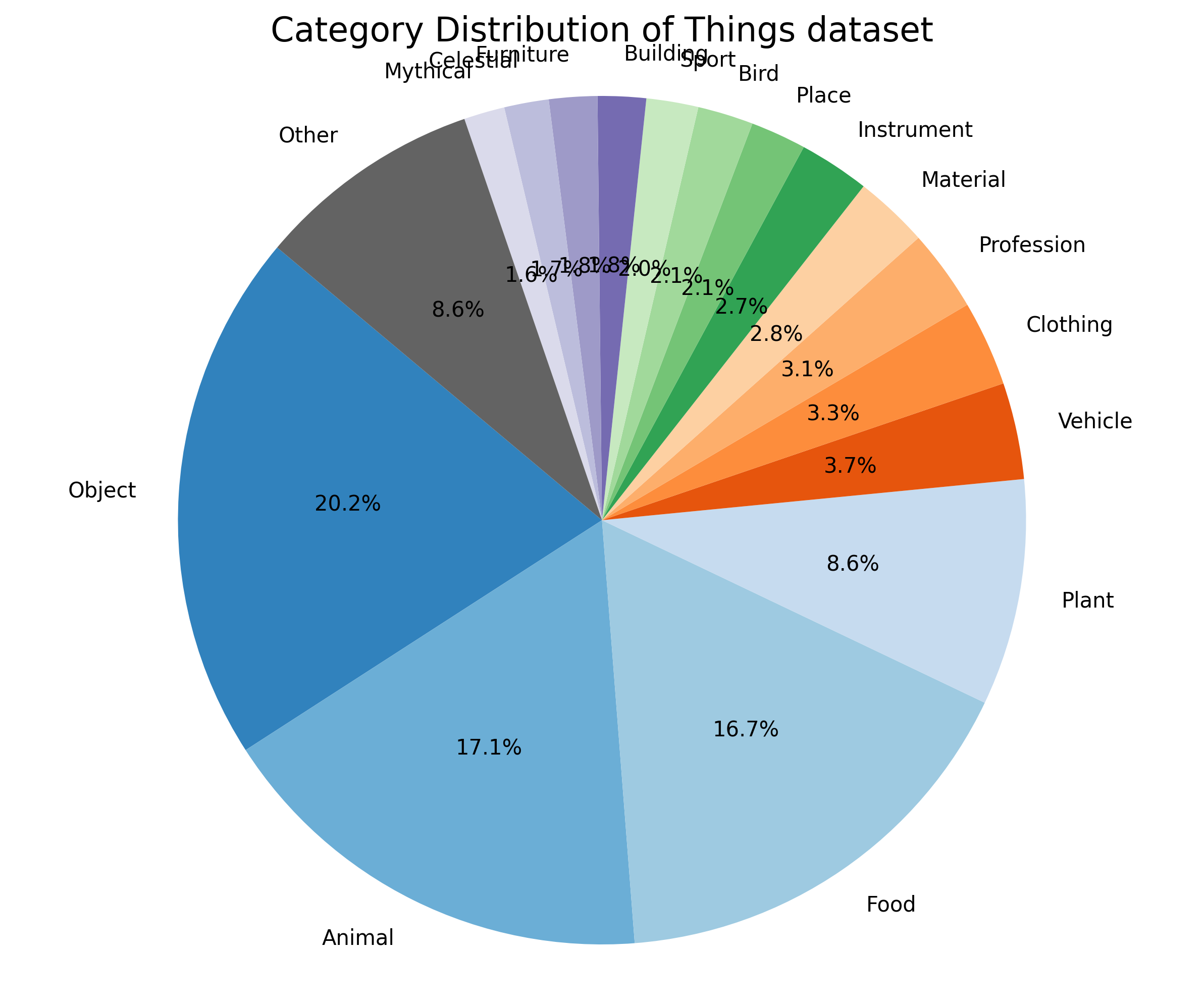}
    \includegraphics[width=1.0\linewidth]{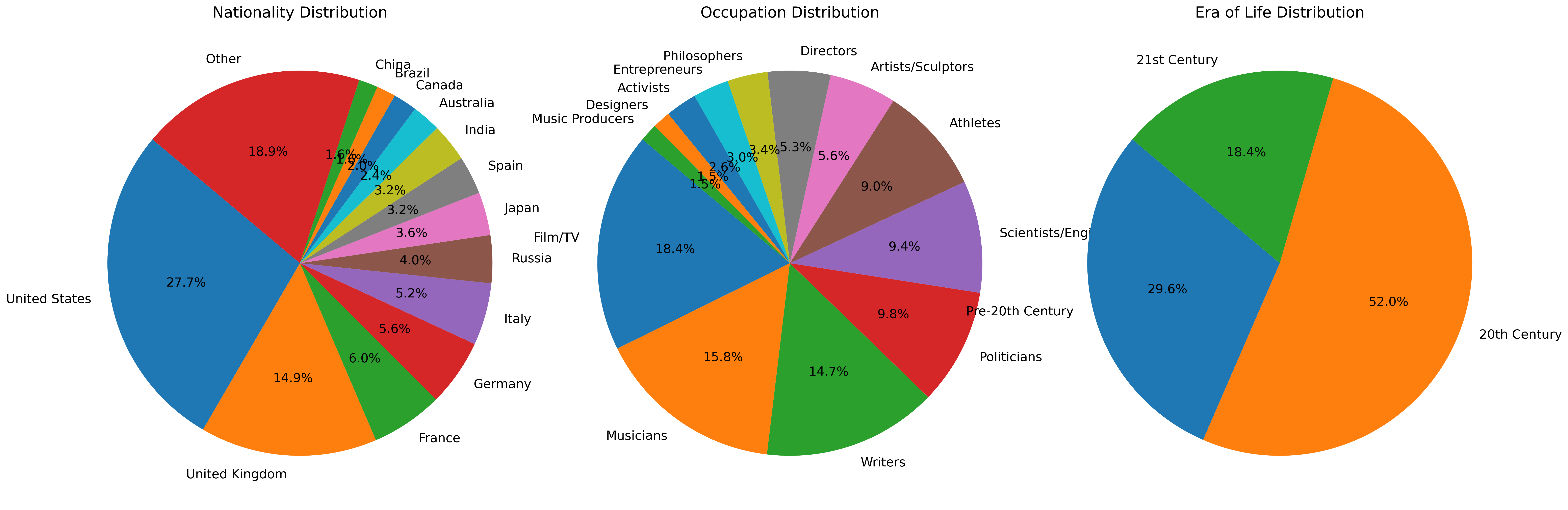}
    \caption{Composition of \ours \textit{Things} and \textit{Celebrities} datasets.
    }
    \label{fig:dataset}
\end{figure}

We provide the dataset composition on \textit{Things} and \textit{Celebrities} in Table~\ref{fig:dataset}.

\section{Prompt used for \textit{Celebrities} and rationale for using ``Dunno''}
\label{app:celeb}
For \textit{Celebrities}, the \texttt{J} use the following prompt to generate a response. 
\begin{center}
\begin{tcolorbox}[width=0.99\textwidth]
\textit{Based on on your knowledge about the celebrity: \textcolor{blue}{\{entity\}}
respond to the following question or guess.
Limit your respond to only 'Yes.', 'No.' or 'Dunno.', with no explanation or other words.
Never say the name \{entity\} in your response. Do not say 'Dunno.' if it can be answered by 'Yes.' or 'No.'
If the question is to solicit the answer, respond 'No.'.\textcolor{gray}{\textbackslash n}
Question/Guess: \textcolor{blue}{\{question\}}(Yes/No/Dunno)}
\end{tcolorbox}
\end{center}

The \texttt{G} use the following prompt to generate a question or guess. 
\begin{center}
\begin{tcolorbox}[width=0.99\textwidth]
\textit{Your task is to ask a series of questions to deduce the celebrity that I'm thinking of with as few queries as possible. Only ask questions that can be answered by ``Yes'', ``No'', or ``Dunno''. Do not ask for hint. Make your question brief with no linebreaker. Now start asking a question.
\textcolor{gray}{\textbackslash n}
\textcolor{blue}{\{dialog history\}}}
\end{tcolorbox}
\end{center}

\paragraph{Rationale for using ``Dunno''}
The Things dataset contains entities that has many different forms, such as if asking ``is this made by metal?'' for the entity ``Chopstick''. The answer to this question could be ``Maybe'' because sometimes they are made by metal and sometimes are not. On the other hand, in the Celebrities dataset, most questions can typically be clearly answered with a yes or no. These questions often refer to gender, birth year, or other specific information about a person. However, there is a possibility that the model might not know the answer to certain questions, such as ``Does this celebrity smoke?''. In these cases, the answer could be ``Dunno'' as it is considered very personal information. As a general observation, we have found that less than 1\% of the responses in the Celebrities dataset are ``Dunno'', while around 20\% of the responses in the Things dataset are ``Maybe''. Note that the percentage of ``Dunno'' or ``Maybe'' responses can vary depending on the specific \texttt{G} being used.

\paragraph{Human baseline}
\label{app:human_baseline}
Collecting static human annotation for this study is a challenging task due to the interactive nature of this research. In order to establish a baseline of human performance, we conducted a human-in-the-loop study. We subsampled \textit{Things} and \textit{Celebrities} into 2 smaller subsets containing 30 sample datapoints each, \textit{Things-30} and \textit{Celebrities-30}. We set up a game server and recruited 108 non-paid internal volunteers to interact with the \texttt{J}, and collected a total of 145 and 71 human game play sessions for \textit{Things-30} and \textit{Celebrities-30}, respectively. Human guessers were given the same instructions as the LLM guessers and were provided with a tutorial. Optionally, for training, they could also view the question GPT-3.5 would have asked at the previous step. Statistics, experimental details and UI are provided in Appendix~\ref{app:UI}.

To ensure the quality of the data, we manually inspected human game plays to filter out sessions that do not complete the entity deduction task, contain irrelavant chit-chat, or contain nonsensical repeated questions, leaving only sensible game plays. In the end 201 sessions (93.1\% of the total game play) were selected as valid game play.

\begin{table}[ht!]
\scriptsize
\centering
\rowcolors{2}{gray!25}{white} %
\begin{tabular}{lcccccccc}
    \toprule
    & \multicolumn{4}{c}{\textbf{Things-30}} & \multicolumn{4}{c}{\textbf{Celebrities-30}} \\
    \cmidrule(r){2-5} \cmidrule(r){6-9}
     & \textbf{\#Turns ($\downarrow$)} & \textbf{Success  ($\uparrow$)} & \textbf{\#Yes} & \textbf{Score ($\uparrow$)} & \textbf{\#Turns ($\downarrow$)} & \textbf{Success ($\uparrow$)} & \textbf{\#Yes} & \textbf{Score ($\uparrow$)} \\
    \midrule
    GPT-4 & \textbf{16.9$\pm$0.2} & \textbf{0.49$\pm$0.06} & 6.0$\pm$0.2 & \scorecolor{20}\textbf{0.40$\pm$0.05} & 16.5$\pm$0.5 &\textbf{0.59$\pm$0.04} & 7.3$\pm$0.1 & \scorecolor{24}\textbf{0.48$\pm$0.03} \\
    GPT-3.5 & 18.4$\pm$0.3 & 0.25$\pm$0.04 & 7.1$\pm$0.4 & \scorecolor{11}0.21$\pm$0.04 & 17.9$\pm$0.3 & 0.41$\pm$0.05 & 7.6$\pm$0.3 & \scorecolor{16}0.33$\pm$0.04 \\
    Claude-2 & 17.6$\pm$0.3 & 0.29$\pm$0.05 & 4.5$\pm$0.3 & \scorecolor{13}0.25$\pm$0.04 & \textbf{15.9$\pm$0.4} & 0.45$\pm$0.06 & 5.3$\pm$0.1 & \scorecolor{20}0.40$\pm$0.05 \\
    Claude-1 & 18.7$\pm$0.1 & 0.15$\pm$0.02 & 4.3$\pm$0.2 & \scorecolor{7}0.13$\pm$0.02 & 16.7$\pm$0.4 & 0.41$\pm$0.05 & 4.6$\pm$0.2 & \scorecolor{18}0.35$\pm$0.04 \\
    Vicuna 13B & 18.7$\pm$0.2 & 0.20$\pm$0.03 & 5.2$\pm$0.3 &\scorecolor{8}0.17$\pm$0.02 & 17.7$\pm$0.4 & 0.36$\pm$0.08 & 6.8$\pm$0.3 &\scorecolor{14}0.27$\pm$0.06 \\
    Vicuna 7B & 19.1$\pm$0.4 & 0.11$\pm$0.06 & 5.7$\pm$0.6 & \scorecolor{5}0.10$\pm$0.05 & 19.7$\pm$0.3 & 0.05$\pm$0.04 & 6.2$\pm$0.7 & \scorecolor{2}0.04$\pm$0.03 \\
    \midrule
    V-FT 7B (All) & 18.4$\pm$0.2 & 0.20$\pm$0.02 & 6.8$\pm$0.2 & \scorecolor{9}0.17$\pm$0.02 & 19.0$\pm$0.2 & 0.21$\pm$0.04 & 9.1$\pm$0.3 & \scorecolor{8}0.16$\pm$0.03 \\
    V-FT 7B (Things) & 18.5$\pm$0.4 & 0.22$\pm$0.06 & 6.6$\pm$0.2 & \scorecolor{9}0.18$\pm$0.05  & 19.1$\pm$1.5 & 0.19$\pm$0.20 & 10.3$\pm$3.6 & \scorecolor{8}0.15$\pm$0.17 \\
    V-FT 7B (Celebs) & 19.7$\pm$0.3 &	0.03$\pm$0.02 & 1.6$\pm$0.1	&\scorecolor{1.5}0.03$\pm$0.02 & 19.1$\pm$0.2 & 0.20 $\pm$0.07 & 7.5$\pm$0.6 & \scorecolor{8}0.16$\pm$0.05 \\
    V-FT 7B (Suc.) & 18.5$\pm$0.5 & 0.28$\pm$0.10 & 6.8$\pm$0.5& \scorecolor{11.5}0.23$\pm$0.08 & 18.6$\pm$0.5 & 0.21$\pm$0.06 & 7.4$\pm$1.4 & \scorecolor{9}0.17$\pm$0.04 \\
    V-FT 13B (Suc.) & 18.0$\pm$0.5 & 0.29$\pm$0.08 & 6.9$\pm$0.2 & \scorecolor{12}0.24$\pm$0.07  & 18.6$\pm$0.6 & 0.22$\pm$0.09 & 7.8$\pm$0.5 & \scorecolor{9}0.18$\pm$0.07 \\
    \midrule
    V-RLGP 7B &19.3$\pm$0.2 & 0.15$\pm$0.03 & 3.6$\pm$0.1 & \scorecolor{6}0.12$\pm$0.02&		19.5$\pm$0.3&	0.09$\pm$0.05&	5.8$\pm$1.1&	\scorecolor{4}0.07$\pm$0.04 \\
    V-RLGP 13B & 17.8$\pm$0.2 & 0.31$\pm$0.03 & 4.0$\pm$0.2 & \scorecolor{13}0.26$\pm$0.02	&	17.5$\pm$0.5	&0.35$\pm$0.04	&6.8$\pm$0.2	&\scorecolor{15}0.29$\pm$0.04 \\
    \midrule
    Human & 18.5$\pm$0.5 &	0.24$\pm$0.04 &	5.2$\pm$0.2 &	\scorecolor{10}0.20$\pm$0.04 & 18.1$\pm$0.2  &	0.31$\pm$0.03 & 7.0$\pm$0.3 &	\scorecolor{12}0.25$\pm$0.03 \\
    \bottomrule
\end{tabular}
\vspace{4mm}
\caption{Benchmark of LLMs on the \ours datasets (\textit{Things-30}, \textit{Celebrities-30}) including human baseline. \textbf{\#Turns} denotes the average number of turns taken. \textbf{Success} denotes the percentage of the games that the \texttt{G} wins. \textbf{\#Yes} denotes the average number of ``yes'' response received from \texttt{J}. \textbf{Score} indicates the reward score defined in ~\cref{eq:score}. 
The mean and standard deviation were calculated based on 5 repetitions.
}
\label{tab:llm_eval_old}
\end{table}

\section{Additional results and prompt used for Dialog State Probing}
\label{app:dialog-state}
We used the following prompt to probe the model for its top guesses during each turn. 
\begin{center}
\begin{tcolorbox}[width=0.99\textwidth]
\textit{
\textcolor{blue}{\{dialog history\}}\\
Based on the information provided, here are the top 5 most likely concrete entities I think you are thinking of:}
\end{tcolorbox}
\end{center}
We provide additional top K probing results from GPT-4 on \textit{Celebrities} in Table~\ref{tab:gpt_celeb_top_k}. Similarly, the model seems to be able to leverage its own hierarchical representation of the celebrities' space to ask questions that steer to narrow down the search effectively. 

\section{Measurement of uncertainty}
\label{app:uncertain}
We asked the Guesser model, ``On a scale of 1-5, how confident are you in your top guesses based on the current conversation history?'' We observed a correlation between the model's predicted uncertainty and significant shifts in its top guesses. However, this pattern was mainly observed in more powerful models such as GPT-4. In the case of weaker models like Vicuna, the top guesses or confidence scores generated by the model were not very sensible.

\section{Full ablation results on RQ2, planning vs reasoning}
\label{app:reason_vs_plan}
\begin{table}[ht!]
\centering
\begin{tabular}{lcccc}
    \toprule
    & \textbf{Things-30 ($\uparrow$)} & \textbf{Celebs-30 ($\uparrow$)} \\
    \midrule
    GPT-4 $\rightarrow$ Vicuna 7B & \scorecolor{6}0.12$\pm$0.03 & \scorecolor{10}0.19$\pm$0.02 \\
    GPT-4 $\rightarrow$ GPT-4 & \scorecolor{20}\textbf{0.40$\pm$0.05} & \scorecolor{24}\textbf{0.48$\pm$0.03} \\
    Vicuna 7B $\rightarrow$ Vicuna 7B & \scorecolor{5}0.10$\pm$0.04 & \scorecolor{2}0.04$\pm$0.03 \\
    Vicuna 7B $\rightarrow$ GPT-4 & \scorecolor{5}0.11$\pm$0.03 & \scorecolor{3}0.06$\pm$0.03 \\
    Human $\rightarrow$ Vicuna 7B & \scorecolor{6}0.11$\pm$0.04 & \scorecolor{2}0.03$\pm$0.01 \\
    Human $\rightarrow$ GPT-4 & \scorecolor{8}0.15$\pm$0.01 & \scorecolor{9}0.18$\pm$0.06 \\
    Human $\rightarrow$ Human & \scorecolor{10}0.20$\pm$0.04 & \scorecolor{13}0.25$\pm$0.03 \\
    \bottomrule
\end{tabular}
\caption{Ablation on planning and reasoning ability. Note that non-human game plays are also evaluated on the subsampled datasets \textit{Things-30} and \textit{Celebrities-30} for a fair comparison. Numbers are the game scores with 5 repetitions. X $\rightarrow$ Y indicates using the conversation trajectory from X and swap the Y for the last turn. For example, GPT-4 $\rightarrow$ Vicuna 7B uses GPT-4 to play all except the last turn, swapping in Vicuna 7B in the last turn. }
\label{tab:planning_reasoning2}
\end{table}

The full results comparing planning and reasoning are presented in Table~\ref{tab:planning_reasoning2}.
Interestingly, when GPT-4 is employed for the last step reasoning in human trajectories, there is a decrease in performance. One possible explanation for this is that while GPT-4 may excel in comprehending model-generated trajectories, it may struggle to fully interpret the planning trajectory of humans based on the observed turns. This suggests that models may be more compatible with model-generated trajectories rather than human trajectories.

\section{Experimental details for human baseline}
\label{app:UI}

In total, we recruited 108  human volunteers to participate in the study. To ensure the quality of the data, we manually inspected human game plays to filter out low quality data.
We also prioritized games that received fewer qualified plays to present to the human players. After a 30-day trial period, we collected a total of 140 and 68 human game play sessions for \textit{Things} and \textit{Celebrities}, respectively. Each entity or celebrity is covered at least by two game plays.

\begin{figure}[ht!]
    \centering
    \begin{minipage}{0.48\linewidth}
        \centering
        \includegraphics[width=1.0\linewidth]{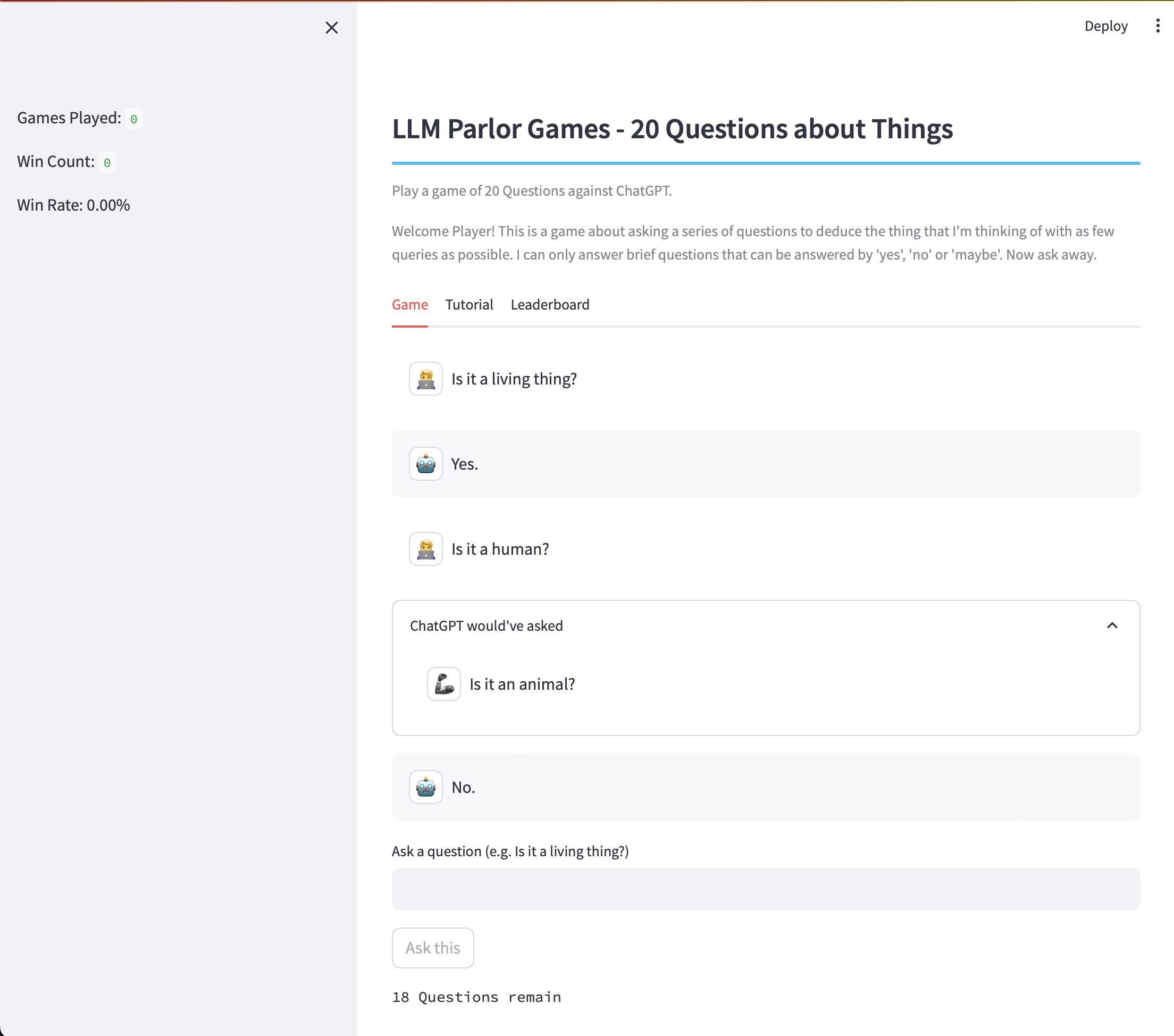}
    \end{minipage}\hfill
    \begin{minipage}{0.48\linewidth}
        \centering
        \includegraphics[width=1.0\linewidth]{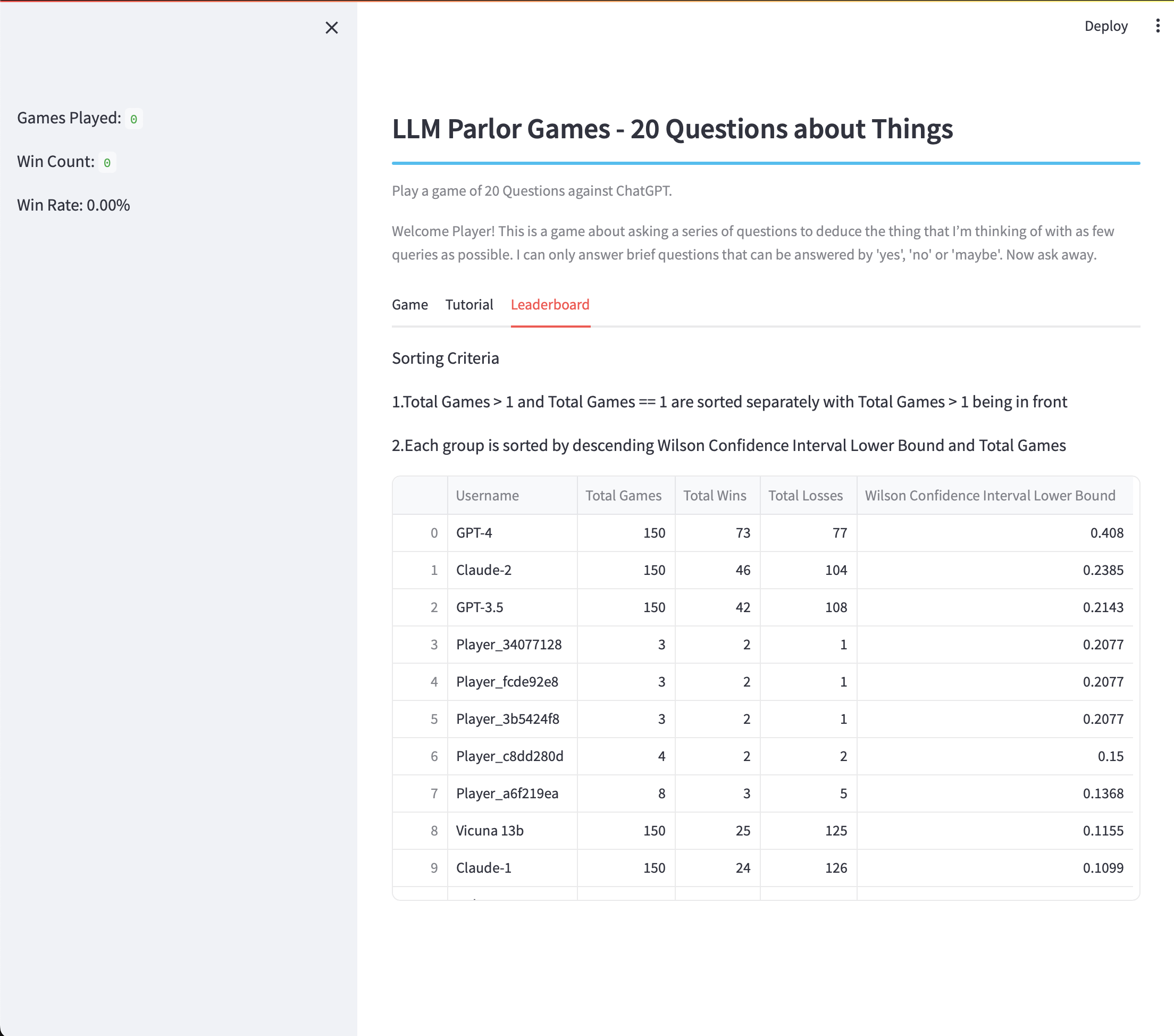}
    \end{minipage}\hfill
    \caption{Game play UI interface for collecting human baseline. On the left, human players are given prompt instructions equivalent to LLM guessers. An optional retrospection UI can be toggled to display what ChatGPT would've chosen to ask in the last turn. On the right, a leaderboard with Human and LLM player performance is shown.}
    \label{fig:ui}
\end{figure}

We show the UI for the in game demo server in Figure~\ref{fig:ui}. The UI contains 3 sections -- game play, tutorial and leaderboard. In the leaderboard, we compute the Wilson confidence interval for each player, and include the performance of LLMs as benchmarks for the human players to refer to. 

For Human game plays, entities with fewer game plays are oversampled so that all entities have the same amount of game play. Game plays are then randomized into separate runs to calculate means and standard deviations. Note that this might not be exactly comparable to the model's numbers.

\section{Comparison of GPT-4 vs Human player}
\label{app:GPT}
We provide additional examples in Table~\ref{tab:gptvshuman} (\textit{Things}) and Table~\ref{tab:gptvshuman_celeb} (\textit{Celebrities}) for GPT-4's generation and compare them with the human players, highlighting that strong LLMs share similarities with humans in their ability to \textit{backtrack} when they are on the wrong path. One advantage of LLM over humans is that LLM may have a superior level of knowledge (Table~\ref{tab:gptvshuman_celeb}), empowering them to ask questions using \textit{tacit knowledge} among LLMs.
One weakness of the LLMs comparing to humans is that LLMs tend to enumerate or repeat to follow the same pattern of previous turns, demonstrating \textit{self-reinforced} undesirable behaviors. 
Weaker models tend to be attracted to this absorbing state of repetition or artifacts, and accumulate errors until they ultimately fail (Appendix~\ref{app:failure_mode}). 
Stronger models occasionally display the ability to escape from this repetitive patterns and rectify early errors, which presumably be attributed to their stronger adherence to the high-level plan.
We then present a detailed analysis on each dataset: 

\subsection{\textit{Things} dataset}
In the first example ``Guitar'', we observed that both GPT-4 and the human player efficiently converged on a music-related object within approximately 10 turns. However, the human player deviated from the correct path by thinking about an electronic device, whereas GPT-4 promptly identified the item as a musical instrument. Notwithstanding this, the human player's capacity for back-tracking their steps was evident when they recognized the erroneous direction and ultimately win the game. In this game, the human player retraced the error and asking a reconfirming question ``Is it some kind of music player?''.

We then ask the question if the GPT-4 model also has this back-tracking ability. The answer is yes. In Table~\ref{tab:gptvshuman} (Guitar, Human + GPT-4), we ask GPT-4 to complete the game based on partial gameplay provided by human players until they recognize their mistake (up to ``\texttt{G}: Can it display images? \texttt{J}: No.''). GPT-4 promptly comprehends the game scenario and identifies that it has taken the incorrect route. Subsequently, it formulates the appropriate question that encompasses the overlooked category, thereby salvaging the game. We have also observed this similar capability in several other game sessions from GPT-4.

In the second game ``Bald eagle'', human player has a luck guess after identifying the entity is a bird. GPT-4, on the contrary, still roll out 3 more turns to gather more information before making an informative guess. 

In the third example ``Jigsaw Puzzle'', GPT-4 interleaves questions regarding size,  purpose and location and combine multiple aspects of information to make a guess.  Prior to making the final guess, GPT-4 demonstrates caution by reconfirming if the object is a physical toy, after receiving a ``No'' with ``Is it a board game?''. On the contrary, the human player seems clueless and may ask repetitive or irrelevant questions. 

\subsection{\textit{Celebrities} dataset}
We present a comparative analysis between GPT-4 and human players on \textit{Celebrities} in Table~\ref{tab:gptvshuman_celeb}. Our findings reveal that GPT-4 demonstrates a propensity for asking questions that necessitate domain-specific background knowledge (as highlighted in Table~\ref{tab:gptvshuman_celeb}). Such knowledge could be \textit{tacit knowledge} between LLMs, which human players may lack. 
With its extensive information repository, GPT-4's inquiries can delve into highly specific topics, sometimes requiring professional expertise that human judges may not possess. This characteristic confers an additional strength and advantage to GPT-4 during gameplay. Conversely, human players tend to rely more on direct guesses, which may yield fortunate outcomes in certain scenarios, but limits their overall effectiveness when compared to GPT-4.

\section{Game runs and failure modes on \textit{Celebrities} and \textit{Things}}
\label{app:failure_mode}
We provide additional gameplay sessions on \textit{Celebrities}, as shown in Table~\ref{tab:failure_mode_celeb}. Similar findings can be observed as they are in the experiment section. We observed that stronger models display proficient gameplay by accurately identifying the relevant question that separates the current space, while weaker models encounter difficulties in planning and reasoning.

We also presents several other failure modes observed with the Vicuna 7b model in Table~\ref{tab:artifact}. These failure modes illustrate the repetitive behavior and generation of artifacts. When a less robust model is drawn towards an absorbing state of repetitive pattern, it becomes increasingly challenging for them to escape from it.

\section{Experimental settings for Bahavior Cloning}
\label{app:FT}
We used 4x A100 Nvidia GPU for training. The learning rate was configured to be 2e-5, while the batch size was set to 32. We utilized the BF16 method for training and consistently employed deepspeed stage 2 across all our experiments. As part of the training process, we masked tokens from the turns of the \texttt{J} within the loss function. In all of our experiments, we continued training the model until the validation loss ceased to decrease any further.

\section{Generalization ability of Vicuna 7B finetuned on \textit{Things} to  \textit{Celebrities}}
\label{app:ft_vs_sc}
A comparison of the gameplay between Vicuna 7B and the \textbf{V-FT 7B (Things)} is provided in Table~\ref{tab:ft_vs_sc}. We observe that Vicuna 7B model may be more inclined to ask irrelevant questions, while the V-FT 7B (things) ask questions directly to the point. In terms of game strategy, the V-FT 7B (things) is more generative effective, indicating the planning ability obtained from imitation might be able to generalize to out of domain scenario.  

\section{Experimental settings for Reinforcement Learning from Game-Play (RLGP)}
\label{app:RLGP}
We adhered to the default settings of the TRLX library for our experiments, but made modifications to certain hyperparameters. Specifically, we set the rollout generation temperature to 0.8 and disabled the topK or topP sampling. The value function coefficient for the value hydra head was set to 0.05. For the discounted reward, we set $\gamma$ to 1 and $\lambda$ to 0.97. The RL models were trained using four Nvidia A100 cards, with a total of 32 rollouts for each training entity. We trained the Vicuna 7B and 13B models on a dataset of 209 entities from the \textit{Things} domain, which at least one of the off-the-shelf LLMs had won at least once. This is to ensure that model learn from experiences where it has some chance to receive some final rewards. 

The intermediate reward was determined as follows. We utilized a linear decaying function to gradually decrease the intermediate reward from 2 to 0, based on the following equation:

\begin{align}
    R_{\text{intermediate}} =  \max(0.2 - 0.025 \cdot \textbf{\#Turns}, 0)
\end{align}

The rationale behind this choice is that achieving more ``yes'' responses in the earlier turns is considered a positive signal. We observed that this approach improved the performance of the RLGP model. If the initial intermediate reward is too high (\textit{e.g.,} 0.5), the model tends to generate trivial questions like ``is it an entity'' more frequently. Our choice is based the empirical results obtained from the development set. All RLGP models were trained for 600 iterations, ensuring that the model does not deviate significantly from the initial Vicuna model.

\begin{table}[h]
    \centering
    \scriptsize
    \begin{tabular}{p{1.6in}|p{1.5in}|p{1.5in}}
    \toprule
    \multicolumn{3}{c}{Item: Guitar} \\
    \hline
    \thead{GPT-4} & \thead{Human player} &\thead{Human + GPT-4} \\
    \hline
    \makecell[l]{\texttt{G}: Is it a living entity?  \user{\texttt{J}: No.}\\ \texttt{G}: Is it man-made?  \user{\texttt{J}: Yes.}\\ \texttt{G}: Can it be held in one hand?  \user{\texttt{J}: Yes.}\\ \texttt{G}: Is it electronic?  \user{\texttt{J}: No.}\\ \texttt{G}: Is it used for writing or drawing?  \user{\texttt{J}: No.}\\ \texttt{G}: Is it made of metal?  \user{\texttt{J}: No.}\\ \texttt{G}: Is it used in cooking or eating?  \user{\texttt{J}: No.}\\ \texttt{G}: Is it used for entertainment or leisure?  \\ \quad \user{\texttt{J}: Yes.}\\ \texttt{G}: Is it associated with a specific sport?  \\ \quad \user{\texttt{J}: No.}\\ \texttt{G}: Is it used in board games?  \user{\texttt{J}: No.}\\ \texttt{G}: Is it associated with music?  \user{\texttt{J}: Yes.}\\ \texttt{G}: Is it a musical instrument?  \user{\texttt{J}: Yes.}\\ \texttt{G}: Is it a stringed instrument?  \user{\texttt{J}: Yes.}\\ \texttt{G}: Is it commonly used in a symphony \\ \quad orchestra?  \user{\texttt{J}: No.}\\ \texttt{G}: Is it a guitar?  \user{\texttt{J}: Bingo!}}&
    \makecell[l]{\texttt{G}: Is it a living thing?  \user{\texttt{J}: No.}\\ \texttt{G}: Is it a man-made object?  \user{\texttt{J}: Yes.}\\ \texttt{G}: Is it commonly used indoors?  \user{\texttt{J}: Yes.}\\ \texttt{G}: Is it larger than a toaster?  \user{\texttt{J}: Yes.}\\ \texttt{G}: Is it used for recreational purposes?  \\ \quad \user{\texttt{J}: Yes.}\\ \texttt{G}: Does it require electricity?  \user{\texttt{J}: Maybe.}\\ \texttt{G}: Does it have a screen?  \user{\texttt{J}: No.}\\ \texttt{G}: Does it contain metal?  \user{\texttt{J}: Yes.}\\ \texttt{G}: Does it contain glass?  \user{\texttt{J}: No.}\\ \texttt{G}: Does it make sound?  \user{\texttt{J}: Yes.}\\ \texttt{G}: Is it used to play music?  \user{\texttt{J}: Yes.}\\ \texttt{G}: Is it a speaker?  \user{\texttt{J}: No.}\\ \texttt{G}: Is it a radio?  \user{\texttt{J}: No.}\\ \texttt{G}: Does it have a wireless connection?  \\ \quad \user{\texttt{J}: No.}\\ \texttt{G}: Can it display images?  \user{\texttt{J}: No.}\\ \texttt{G}: \error{Is it some kind of music player?}  \\ \quad \user{\texttt{J}: No.}\\ \texttt{G}: Is it a musical instrument?  \user{\texttt{J}: Yes.}\\ \texttt{G}: Is it a guitar?  \user{\texttt{J}: Bingo!}} &
    \makecell[l]{\texttt{G}: Is it a living thing?  \user{\texttt{J}: No.}\\ \texttt{G}: Is it a man-made object?  \user{\texttt{J}: Yes.}\\ \texttt{G}: Is it commonly used indoors?  \user{\texttt{J}: Yes.}\\ \texttt{G}: Is it larger than a toaster?  \user{\texttt{J}: Yes.}\\ \texttt{G}: Is it used for recreational purposes?  \\ \quad \user{\texttt{J}: Yes.}\\ \texttt{G}: Does it require electricity?  \user{\texttt{J}: Maybe.}\\ \texttt{G}: Does it have a screen?  \user{\texttt{J}: No.}\\ \texttt{G}: Does it contain metal?  \user{\texttt{J}: Yes.}\\ \texttt{G}: Does it contain glass?  \user{\texttt{J}: No.}\\ \texttt{G}: Does it make sound?  \user{\texttt{J}: Yes.}\\ \texttt{G}: Is it used to play music?  \user{\texttt{J}: Yes.}\\ \texttt{G}: Is it a speaker?  \user{\texttt{J}: No.}\\ \texttt{G}: Is it a radio?  \user{\texttt{J}: No.}\\ \texttt{G}: Does it have a wireless connection?  \\ \quad \user{\texttt{J}: No.}\\ \texttt{G}: Can it display images?  \user{\texttt{J}: No.}\\ \texttt{GPT}: \textcolor{blue}{Is it a musical instrument?}  \\ \quad \user{\texttt{J}: Yes.}\\ \texttt{GPT}: \textcolor{blue}{Is it a string instrument?}   \user{\texttt{J}: Yes.}\\ \texttt{GPT}: \textcolor{blue}{Is it a guitar?}  \user{\texttt{J}: Bingo!}} 
    \\
    \end{tabular}
    \begin{tabular}{p{2.3in}|p{2.3in}}
    \toprule
    \multicolumn{2}{c}{Item: Bald eagle} \\
    \hline
    \thead{GPT-4} & \thead{Human player} \\
    \hline
    \makecell[l]{\texttt{G}: Is it a living thing?  \user{\texttt{J}: Yes.}\\ \texttt{G}: Is it an animal?  \user{\texttt{J}: Yes.}\\ \texttt{G}: Is it a domestic animal?  \user{\texttt{J}: No.}\\ \texttt{G}: Is it a mammal?  \user{\texttt{J}: No.}\\ \texttt{G}: Is it a reptile?  \user{\texttt{J}: No.}\\ \texttt{G}: Is it a bird?  \user{\texttt{J}: Yes.}\\ \texttt{G}: Is it a bird of prey?  \user{\texttt{J}: Yes.}\\ \texttt{G}: Does it have a predominantly white color?  \user{\texttt{J}: Yes.}\\ \texttt{G}: Is it found in the Polar regions?  \user{\texttt{J}: No.}\\ \texttt{G}: Is it a bald eagle?  \user{\texttt{J}: Bingo!}}&
    \makecell[l]{\texttt{G}: Is it alive?  \user{\texttt{J}: Yes.}\\ \texttt{G}: Does it live on land?  \user{\texttt{J}: Yes.}\\ \texttt{G}: Is it a plant?  \user{\texttt{J}: No.}\\ \texttt{G}: Is it human?  \user{\texttt{J}: No.}\\ \texttt{G}: Does it live in the Americas?  \user{\texttt{J}: Yes.}\\ \texttt{G}: Does it live in North America?  \user{\texttt{J}: Yes.}\\ \texttt{G}: Is it a predator?  \user{\texttt{J}: Yes.}\\ \texttt{G}: Is it a bear?  \user{\texttt{J}: No.}\\ \texttt{G}: Is it a bird?  \user{\texttt{J}: Yes.}\\ \texttt{G}: Is it a bald eagle?  \user{\texttt{J}: Bingo!}} \\
    \toprule
    \multicolumn{2}{c}{Item: Jigsaw Puzzle} \\
    \hline
    \thead{GPT-4} & \thead{Human player} \\
    \hline
    \makecell[l]{\texttt{G}: Is it a living entity?  \user{\texttt{J}: No.}\\ \texttt{G}: Is it man-made?  \user{\texttt{J}: Yes.}\\ \texttt{G}: Is it used for communication?  \user{\texttt{J}: No.}\\ \texttt{G}: Is it larger than a microwave?  \user{\texttt{J}: Maybe.}\\ \texttt{G}: Is it used for transportation?  \user{\texttt{J}: No.}\\ \texttt{G}: Is it made of metal?  \user{\texttt{J}: No.}\\ \texttt{G}: Is it used in homes?  \user{\texttt{J}: Yes.}\\ \texttt{G}: Is it used for sitting?  \user{\texttt{J}: No.}\\ \texttt{G}: Is it used for storage?  \user{\texttt{J}: No.}\\ \texttt{G}: Is it a decorative item?  \user{\texttt{J}: Maybe.}\\ \texttt{G}: Is it used in the kitchen?  \user{\texttt{J}: No.}\\ \texttt{G}: Is it used in the living room?  \user{\texttt{J}: Maybe.}\\ \texttt{G}: Is it used for entertainment purposes?  \user{\texttt{J}: Yes.}\\ \texttt{G}: Is it a musical instrument?  \user{\texttt{J}: No.}\\ \texttt{G}: Is it a television?  \user{\texttt{J}: No.}\\ \texttt{G}: Is it a game or a toy?  \user{\texttt{J}: Yes.}\\ \texttt{G}: Is it a board game?  \user{\texttt{J}: No.}\\ \texttt{G}: Is it a video game console?  \user{\texttt{J}: No.}\\ \texttt{G}: Is it a physical toy?  \user{\texttt{J}: Yes. You must guess now, what's it?}\\ \texttt{G}: Is it a puzzle? Is it right?  \user{\texttt{J}: Yes.}}&
    \makecell[l]{\texttt{G}: What is the meaning of life?  \user{\texttt{J}: No.}\\ \texttt{G}: What is a good thing?  \user{\texttt{J}: Maybe.}\\ \texttt{G}: Is it human?  \user{\texttt{J}: No.}\\ \texttt{G}: Is it an object?  \user{\texttt{J}: Yes.}\\ \texttt{G}: Is it a tool?  \user{\texttt{J}: Yes.}\\ \texttt{G}: Is it a hammer?  \user{\texttt{J}: No.}\\ \texttt{G}: Is it a kitchen tool?  \user{\texttt{J}: No.}\\ \texttt{G}: What is it?  \user{\texttt{J}: Maybe.}\\ \texttt{G}: Does it cut?  \user{\texttt{J}: No.}\\ \texttt{G}: Is it for food?  \user{\texttt{J}: No.}\\ \texttt{G}: Is it for cars?  \user{\texttt{J}: No.}\\ \texttt{G}: Is it big?  \user{\texttt{J}: Maybe.}\\ \texttt{G}: Can you eat it?  \user{\texttt{J}: No.}\\ \texttt{G}: Is it red?  \user{\texttt{J}: No.}\\ \texttt{G}: Is it metal?  \user{\texttt{J}: No.}\\ \texttt{G}: Is it used for communication?  \user{\texttt{J}: No.}\\ \texttt{G}: Is it used for storing?  \user{\texttt{J}: No.}\\ \texttt{G}: Is it used for helping humans?  \user{\texttt{J}: Maybe.}\\ \texttt{G}: Is it used for taking pictures?  \user{\texttt{J}: No.}\\ \texttt{G}: Alarm clock  \user{\texttt{J}: No.}} \\
    \bottomrule
    \end{tabular}
    \caption{The gameplay scenario generated using GPT-4 model vs the gameplay scenario by human players, on \textit{Things}.}
    \label{tab:gptvshuman}
\end{table}

\begin{table}[h]
    \centering
    \scriptsize
    \begin{tabular}{p{2.3in}|p{2.3in}}
    \toprule
    \multicolumn{2}{c}{Celebrity: Reese Witherspoon} \\
    \hline
    \thead{GPT-4} & \thead{Human player} \\
    \hline
    \makecell[l]{\texttt{G}: Is the celebrity you're thinking of alive?  \user{\texttt{J}: Yes.}\\ \texttt{G}: Is this celebrity a man?  \user{\texttt{J}: No.}\\ \texttt{G}: Is this celebrity an actress?  \user{\texttt{J}: Yes.}\\ \texttt{G}: Is she American?  \user{\texttt{J}: Yes.}\\ \texttt{G}: Has she won an Oscar?  \user{\texttt{J}: Yes.}\\ \texttt{G}: Is she over 50 years old?  \user{\texttt{J}: No.}\\ \texttt{G}: Has she starred in a superhero movie?  \user{\texttt{J}: No.}\\ \texttt{G}: Is she blonde?  \user{\texttt{J}: Yes.}\\ \texttt{G}: \error{Has she appeared in a film directed by Martin }  \\ \quad Scorsese? \user{\texttt{J}: No.}\\ \texttt{G}: Has she won her Oscar within the last 10 years?  \user{\texttt{J}: No.}\\ \texttt{G}: Is she known for romantic comedy films?  \user{\texttt{J}: Yes.}\\ \texttt{G}: Is it Reese Witherspoon?  \user{\texttt{J}: Bingo!}}&
    \makecell[l]{\texttt{G}: Male?  \user{\texttt{J}: No.}\\ \texttt{G}: White?  \user{\texttt{J}: Yes.}\\ \texttt{G}: American  \user{\texttt{J}: Yes.}\\ \texttt{G}: American  \user{\texttt{J}: Yes.}\\ \texttt{G}: Actor  \user{\texttt{J}: Yes.}\\ \texttt{G}: Alive  \user{\texttt{J}: Yes.}\\ \texttt{G}: Action movies  \user{\texttt{J}: No.}\\ \texttt{G}: Romantic comedies  \user{\texttt{J}: Yes.}\\ \texttt{G}: Sandra bullock  \user{\texttt{J}: No.}\\ \texttt{G}: Over 50 years old  \user{\texttt{J}: No.}\\ \texttt{G}: Emma stone  \user{\texttt{J}: Dunno.}\\ \texttt{G}: Is the person female?  \user{\texttt{J}: Yes.}\\ \texttt{G}: Meg Ryan  \user{\texttt{J}: No.}\\ \texttt{G}: Amy Schumer  \user{\texttt{J}: No.}\\ \texttt{G}: Kate Hudson  \user{\texttt{J}: No.}\\ \texttt{G}: Julia Roberts  \user{\texttt{J}: No.}\\ \texttt{G}: Cameron Diaz  \user{\texttt{J}: No.}\\ \texttt{G}: Reese Witherspoon  \user{\texttt{J}: Bingo!}} \\
    \toprule
    \multicolumn{2}{c}{Celebrity: Trevor Noah} \\
    \hline
    \thead{GPT-4} & \thead{Human player} \\
    \hline
    \makecell[l]{\texttt{G}: Is the celebrity you're thinking of alive?  \user{\texttt{J}: Yes.}\\ \texttt{G}: Is the celebrity an actor?  \user{\texttt{J}: Yes.}\\ \texttt{G}: Is the celebrity American?  \user{\texttt{J}: No.}\\ \texttt{G}: Is the celebrity male?  \user{\texttt{J}: Yes.}\\ \texttt{G}: Has the celebrity won an Oscar?  \user{\texttt{J}: No.}\\ \texttt{G}: Is the celebrity known for roles in action films?  \user{\texttt{J}: No.}\\ \texttt{G}: Is the celebrity from the United Kingdom?  \user{\texttt{J}: No.}\\ \texttt{G}: Is the celebrity from Australia?  \user{\texttt{J}: No.}\\ \texttt{G}: Is the celebrity from Asia?  \user{\texttt{J}: No.}\\ \texttt{G}: Is the celebrity from Europe?  \user{\texttt{J}: No.}\\ \texttt{G}: Is the celebrity from Canada?  \user{\texttt{J}: No.}\\ \texttt{G}: Is the celebrity from South America?  \user{\texttt{J}: No.}\\ \texttt{G}: \error{Is the celebrity from Africa?}  \user{\texttt{J}: Yes.}\\ \texttt{G}: Is the celebrity known for roles in comedic films?  \user{\texttt{J}: Yes.}\\ \texttt{G}: Is the celebrity Trevor Noah?  \user{\texttt{J}: Bingo!}}&
    \makecell[l]{\texttt{G}: Is the person male?  \user{\texttt{J}: Yes.}\\ \texttt{G}: Is the person on TV?  \user{\texttt{J}: Yes.}\\ \texttt{G}: Does the person host a late night show?  \user{\texttt{J}: Yes.}\\ \texttt{G}: Is the show factual?  \user{\texttt{J}: Dunno.}\\ \texttt{G}: Is the person Stephen Colbert  \user{\texttt{J}: No.}\\ \texttt{G}: Is the person a news anchor?  \user{\texttt{J}: No.}\\ \texttt{G}: Is the person a comedian?  \user{\texttt{J}: Yes.}\\ \texttt{G}: Is the person John Oliver?  \user{\texttt{J}: No.}\\ \texttt{G}: Is the person Trevor Noah?  \user{\texttt{J}: Bingo!}} \\
    \hline
    \makecell[l]{\texttt{G}: Is your celebrity alive?  \user{\texttt{J}: Yes.}\\ \texttt{G}: Is your celebrity an actor?  \user{\texttt{J}: Yes.}\\ \texttt{G}: Is your celebrity female?  \user{\texttt{J}: No.}\\ \texttt{G}: Has your celebrity won an Oscar?  \user{\texttt{J}: No.}\\ \texttt{G}: Is your celebrity American?  \user{\texttt{J}: No.}\\ \texttt{G}: Is your celebrity British?  \user{\texttt{J}: No.}\\ \texttt{G}: Is your celebrity from Australia?  \user{\texttt{J}: No.}\\ \texttt{G}: Is your celebrity from Asia?  \user{\texttt{J}: No.}\\ \texttt{G}: Is your celebrity from Europe?  \user{\texttt{J}: No.}\\ \texttt{G}: Is your celebrity from Canada?  \user{\texttt{J}: No.}\\ \texttt{G}: Is your celebrity from Africa?  \user{\texttt{J}: Yes.}\\ \texttt{G}: Has your celebrity starred in a superhero movie?  \user{\texttt{J}: No.}\\ \texttt{G}: Is your celebrity over the age of 50?  \user{\texttt{J}: No.}\\ \texttt{G}: \error{Has your celebrity acted in a film that was nominated} \\ \quad \error{for an Academy Award? } \user{\texttt{J}: Yes.}\\ \texttt{G}: Is your celebrity from South Africa?  \user{\texttt{J}: Yes.}\\ \texttt{G}: Is your celebrity Trevor Noah?  \user{\texttt{J}: Bingo!}}&
    \makecell[l]{\texttt{G}: Is your celebrity a movie star?  \user{\texttt{J}: Yes.}\\ \texttt{G}: Is the celebrity female?  \user{\texttt{J}: No.}\\ \texttt{G}: Is the celebrity over 50 years old?  \user{\texttt{J}: No.}\\ \texttt{G}: Is the celebrity over 30 years old?  \user{\texttt{J}: Yes.}\\ \texttt{G}: Has this celebrity been in a Marvel movie?  \user{\texttt{J}: No.}\\ \texttt{G}: Is this celebrity known for comedies?  \user{\texttt{J}: Yes.}\\ \texttt{G}: Is this celebrity african american?  \user{\texttt{J}: No.}\\ \texttt{G}: Has this celebrity also been in a TV show?  \user{\texttt{J}: Yes.}\\ \texttt{G}: Has this celebrity been in a movie grossing more than  \\ \quad a billion dollars?  \user{\texttt{J}: No.}\\ \texttt{G}: Is this celebrity still making hit movies?  \user{\texttt{J}: No.}\\ \texttt{G}: Was this celebrity famous in the 90s?  \user{\texttt{J}: No.}\\ \texttt{G}: Was this celebrity famous in the 2000s?  \user{\texttt{J}: Yes.}\\ \texttt{G}: Has this celebrity worked with Judd Apatow?  \user{\texttt{J}: No.}\\ \texttt{G}: Was this celebrity in one of the Hangover movies?  \user{\texttt{J}: No.}\\ \texttt{G}: Is this celebrity American?  \user{\texttt{J}: Yes.}\\ \texttt{G}: Was this person on network TV?  \user{\texttt{J}: Yes.}\\ \texttt{G}: Does this celebrity have a famous spouse?  \user{\texttt{J}: No.}\\ \texttt{G}: Is this celebrity over 40 years old?  \user{\texttt{J}: No.}\\ \texttt{G}: Is it Zac Efron?  \user{\texttt{J}: No.}\\ \texttt{G}: Is it Zachary Levi?  \user{\texttt{J}: No.}} \\
    \bottomrule
    \end{tabular}
    \caption{The gameplay scenario generated using GPT-4 model vs the gameplay scenario by human players, on \textit{Things}. The questions need in-domain knowledge is \error{highlighted}.}
    \label{tab:gptvshuman_celeb}
\end{table}

\begin{table}[h]
    \centering
    \scriptsize
    \begin{tabular}{p{2.3in}|p{2.3in}}
    \toprule
    \thead{Successful game} & \thead{Planning issue (Early Enumeration)} \\
    \hline
    \makecell[l]{\texttt{G}: Is the celebrity alive?  \user{\texttt{J}: Yes.}\\ \texttt{G}: Is the celebrity a male?  \user{\texttt{J}: No.}\\ \texttt{G}: Is she an actress?  \user{\texttt{J}: Yes.}\\ \texttt{G}: Is she American?  \user{\texttt{J}: Yes.}\\ \texttt{G}: Has she won an Academy Award?  \user{\texttt{J}: Yes.}\\ \texttt{G}: Is she above 50 years old?  \user{\texttt{J}: No.}\\ \texttt{G}: Has she starred in a superhero movie?  \user{\texttt{J}: Yes.}\\ \texttt{G}: Is she in the Marvel Cinematic Universe?  \user{\texttt{J}: Yes.}\\ \texttt{G}: Did she play a character that has their own standalone \\  \quad  film in the Marvel Cinematic Universe?  \user{\texttt{J}: Yes.}\\ \texttt{G}: Has she played a human with supernatural powers?  \user{\texttt{J}: No.}\\ \texttt{G}: Did she play the character of Black Widow?  \user{\texttt{J}: No.}\\ \texttt{G}: Did she play Captain Marvel?  \user{\texttt{J}: No.}\\ \texttt{G}: Did she play the character of Wasp in the Ant-Man \\ \quad series?  \user{\texttt{J}: Yes.}\\ \texttt{G}: Is the celebrity Evangeline Lilly?  \user{\texttt{J}: No.}\\ \texttt{G}: Did she play the character of Gamora in the Guardians \\ \quad of the Galaxy series?  \user{\texttt{J}: No.}\\ \texttt{G}: Did she play the character of Pepper Potts in the Iron Man \\ \quad series?  \user{\texttt{J}: Yes.}\\ \texttt{G}: Is the celebrity Gwyneth Paltrow?  \user{\texttt{J}: \textcolor{blue}{Bingo!}}}&
    \makecell[l]{\texttt{G}: Is the celebrity a person who has been in the public eye \\ \quad for at least 10 years?  \user{\texttt{J}: Yes.}\\ \texttt{G}: Is the celebrity primarily known for acting?  \user{\texttt{J}: Yes.}\\ \texttt{G}: Has the celebrity won an Academy Award?  \user{\texttt{J}: No.}\\ \texttt{G}: Has the celebrity been in any Marvel Cinematic Universe \\ \quad films?  \user{\texttt{J}: Yes.}\\ \texttt{G}: Is the celebrity primarily known for their work in the \\ \quad \error{Marvel Cinematic Universe}?  \user{\texttt{J}: No.}\\ \texttt{G}: Has the celebrity been in any \error{DC Extended Universe} \\ \quad \error{films}?  \user{\texttt{J}: No.}\\ \texttt{G}: Has the celebrity been in any \error{Star Wars films}?  \user{\texttt{J}: No.}\\ \texttt{G}: Is the celebrity primarily known for their work in \\ \quad television?  \user{\texttt{J}: No.}\\ \texttt{G}: Has the celebrity been in any reality TV shows?  \user{\texttt{J}: No.} \\ ...} \\
    \toprule
    \thead{Planning issue (Redundancy)} & \thead{Reasoning issue (Inconsistency)} \\
    \hline
    \makecell[l]{\texttt{G}: Is the celebrity a musician?  \user{\texttt{J}: No.}\\ \texttt{G}: Is the celebrity primarily known for their acting career? \\ \quad \user{\texttt{J}: Yes.}\\ \texttt{G}: Is the celebrity a former child actor?  \user{\texttt{J}: No.}\\ \texttt{G}: Is the celebrity primarily known for their work in film? \\ \quad \user{\texttt{J}: Yes.}\\ \texttt{G}: Is the celebrity mostly known for their work in the past  \\ \quad decade?  \user{\texttt{J}: Yes.}\\ \texttt{G}: Is the celebrity a \error{British national}?  \user{\texttt{J}: No.}\\ \texttt{G}: Is the celebrity primarily known for their work in \\ \quad American cinema?  \user{\texttt{J}: Yes.}\\ \texttt{G}: Is the celebrity a former child star who later became a \\ \quad Hollywood movie star?  \user{\texttt{J}: Yes.}\\ \texttt{G}: Is the celebrity Emma Watson?  \user{\texttt{J}: No.}\\ \texttt{G}: Is the celebrity a \error{male actor}?  \user{\texttt{J}: No.}\\ \texttt{G}: Is the celebrity a \error{British female actor}?  \user{\texttt{J}: No.} \\ ...}&
    \makecell[l]{\texttt{G}: Is the celebrity \error{male}?  \user{\texttt{J}: Yes.}\\ \texttt{G}: Has the celebrity won an Academy Award?  \user{\texttt{J}: No.}\\ \texttt{G}: Is the celebrity primarily known for their work in film? \\ \quad  \user{\texttt{J}: No.}\\ \texttt{G}: Is the celebrity primarily known for their work in \\ \quad  television?  \user{\texttt{J}: No.}\\ \texttt{G}: Is the celebrity known for their work in music?  \user{\texttt{J}: No.}\\ \texttt{G}: Does the celebrity have a significant presence on social \\ \quad  media?  \user{\texttt{J}: Yes.}\\ \texttt{G}: Is the celebrity a comedian?  \user{\texttt{J}: No.}\\ \texttt{G}: Is the celebrity a singer?  \user{\texttt{J}: No.}\\ \texttt{G}: Is the celebrity a dancer?  \user{\texttt{J}: No.}\\...\\ \texttt{G}: Is the celebrity a political figure? \\ \quad   \user{\texttt{J}: No. You must guess now}\\ \texttt{G}: The celebrity is \error{Ellen DeGeneres}. Is it right?  \user{\texttt{J}: No.}} \\
    \bottomrule
    \end{tabular}
    \caption{Demonstration of failure modes of the gameplay with hidden celebrity names. The successful gameplay scenario is generated using GPT-4 model as \texttt{G}. The remaining gameplay scenarios are generated using vicuna-7b model as \texttt{G}.}
    \label{tab:failure_mode_celeb}
\end{table}

\begin{table}[h]
    \centering
    \scriptsize
    \begin{tabular}{p{2.3in}|p{2.3in}}
    \toprule
    \thead{Artifact: Chain Reaction} & \thead{Artifact: Self-answering} \\
    \hline
    \makecell[l]{\texttt{G}: Is the entity a living thing?  \user{\texttt{J}: Yes.}\\ \texttt{G}: Is the living thing an animal?  \user{\texttt{J}: Yes.}\\ \texttt{G}: Is the animal a mammal?  \user{\texttt{J}: Yes.}\\ \texttt{G}: Is the mammal a warm-blooded?  \user{\texttt{J}: Yes.}\\ \texttt{G}: Is the warm-blooded mammal a primate?  \user{\texttt{J}: No.}\\ \texttt{G}: Is the primate a marmoset?  \user{\texttt{J}: No.}\\ \texttt{G}: Is the marmoset a rodent?  \user{\texttt{J}: No.}\\ \texttt{G}: Is the rodent a squirrel?  \user{\texttt{J}: No.}\\ \texttt{G}: Is the squirrel a rabbit?  \user{\texttt{J}: No.}}&
    \makecell[l]{\texttt{G}: Is the entity you are thinking of a person? Yes  \user{\texttt{J}: No.}\\ \texttt{G}: Is the entity you are thinking of an animal? Yes  \user{\texttt{J}: Yes.}\\ \texttt{G}: Is the animal you are thinking of a mammal? Yes  \user{\texttt{J}: No.}\\ \texttt{G}: Is the animal you are thinking of a bird? Yes  \user{\texttt{J}: No.}\\ \texttt{G}: Is the animal you are thinking of a reptile? Maybe  \user{\texttt{J}: No.}\\ \texttt{G}: Is the animal you are thinking of an amphibian? Maybe  \user{\texttt{J}: No.}\\ \texttt{G}: Is the animal you are thinking of a fish? Yes  \user{\texttt{J}: No.}\\ \texttt{G}: Is the animal you are thinking of a crustacean? Maybe  \user{\texttt{J}: No.}\\ \texttt{G}: Is the animal you are thinking of an arachnid? Maybe  \user{\texttt{J}: No.}\\ \texttt{G}: Is the animal you are thinking of an insect? Yes  \user{\texttt{J}: No.}\\ \texttt{G}: Is the animal you are thinking of a mollusk? No  \user{\texttt{J}: No.} \\ ...} \\
    \bottomrule
    \end{tabular}
    \caption{Demonstration of failure modes from Vicuna-7b model where the model was trapped by repetitive or problematic patterns.}
    \label{tab:artifact}
\end{table}

\begin{table}[h]
    \centering
    \scriptsize
    \begin{tabular}{p{2.3in}|p{2.3in}}
    \toprule
    \multicolumn{2}{c}{Celebrity: LeBron James} \\
    \hline
    \thead{Vicuna 7B} & \thead{V-FT 7B (Things)} \\
    \hline
    \makecell[l]{\texttt{G}: Was this celebrity a singer?  \user{\texttt{J}: No.}\\ \texttt{G}: Was this celebrity a musician?  \user{\texttt{J}: No.}\\ \texttt{G}: Was this celebrity an actor?  \user{\texttt{J}: No.}\\ \texttt{G}: Was this celebrity a comedian?  \user{\texttt{J}: No.}\\ \texttt{G}: Was this celebrity a public figure?  \user{\texttt{J}: Yes.}\\ \texttt{G}: Was this public figure a politician?  \user{\texttt{J}: No.}\\ \texttt{G}: Was this public figure a sports person?  \user{\texttt{J}: Yes.}\\ \texttt{G}: Was this sports person a professional athlete?  \user{\texttt{J}: Yes.}\\ \texttt{G}: Was this professional athlete a tennis player?  \user{\texttt{J}: No.}\\ \texttt{G}: Was this professional athlete a boxer?  \user{\texttt{J}: No.}\\ \texttt{G}: Was this professional athlete a golfer?  \user{\texttt{J}: No.}\\ \texttt{G}: Was this professional athlete a football player?  \user{\texttt{J}: No.}\\ \texttt{G}: Was this professional athlete a basketball player?  \user{\texttt{J}: Yes.}\\ \texttt{G}: Was this basketball player a member of the NBA?  \user{\texttt{J}: Yes.}\\ \texttt{G}: Was this basketball player active during the 1990s?  \user{\texttt{J}: No.}\\ \texttt{G}: Was this basketball player active during the 2000s?  \user{\texttt{J}: Yes.}\\ \texttt{G}: Was this basketball player known for his clutch  \\ \quad   performances?  \user{\texttt{J}: Yes.}\\ \texttt{G}: Can you name the basketball player?  \user{\texttt{J}: No.}\\ \texttt{G}: I'm sorry, I'll have to ask for a hint. Could it be someone  \\ \quad   who played for the Los Angeles Lakers during the 2000s?  \\ \quad   \user{\texttt{J}: No. You must guess now, what's it?}\\ \texttt{G}: Could it be someone who played for the Los Angeles  \\ \quad   Clippers during the 2000s? Is it right?  \user{\texttt{J}: No.}}&
    \makecell[l]{\texttt{G}: Is the celebrity you're thinking of a male?  \user{\texttt{J}: Yes.}\\ \texttt{G}: Is the celebrity you're thinking of a singer?  \user{\texttt{J}: No.}\\ \texttt{G}: Is the celebrity you're thinking of an actor?  \user{\texttt{J}: No.}\\ \texttt{G}: Is the celebrity you're thinking of a sports player?  \\ \quad   \user{\texttt{J}: Yes.}\\ \texttt{G}: Is the celebrity you're thinking of a male sports player?  \\ \quad   \user{\texttt{J}: Yes.}\\ \texttt{G}: Is the celebrity you're thinking of an Olympic athlete?  \\ \quad   \user{\texttt{J}: Yes.}\\ \texttt{G}: Is the celebrity you're thinking of a swimmer?  \user{\texttt{J}: No.}\\ \texttt{G}: Is the celebrity you're thinking of a track and field athlete?  \\ \quad  \user{\texttt{J}: No.}\\ \texttt{G}: Is the celebrity you're thinking of a gymnast?  \user{\texttt{J}: No.}\\ \texttt{G}: Is the celebrity you're thinking of a boxer?  \user{\texttt{J}: No.}\\ \texttt{G}: Is the celebrity you're thinking of a soccer player?  \user{\texttt{J}: No.}\\ \texttt{G}: Is the celebrity you're thinking of a basketball player?  \\ \quad   \user{\texttt{J}: Yes.}\\ \texttt{G}: Is the celebrity you're thinking of a former NBA player?  \\ \quad    \user{\texttt{J}: Yes.}\\ \texttt{G}: Was the celebrity you're thinking of a dominant scorer?  \\ \quad   \user{\texttt{J}: Yes.}\\ \texttt{G}: Did the celebrity you're thinking of have a reputation for   \\ \quad  being a high-flyer?  \user{\texttt{J}: Yes.}\\ \texttt{G}: Is the celebrity you're thinking of known for their acrobatic   \\ \quad   dunks?  \user{\texttt{J}: Yes.}\\ \texttt{G}: Is the celebrity you're thinking of Michael Jordan?  \user{\texttt{J}: No.}\\ \texttt{G}: Is the celebrity you're thinking of Kobe Bryant?  \user{\texttt{J}: No.}\\ \texttt{G}: Is the celebrity you're thinking of LeBron James?   \\ \quad   \user{\texttt{J}: Bingo!}} \\
    \toprule
    \multicolumn{2}{c}{Celebrity: Brie Larson} \\
    \hline
    \thead{Vicuna 7B} & \thead{V-FT 7B (Things)} \\
    \hline
    \makecell[l]{\texttt{G}: Is the celebrity you're thinking of a male?  \user{\texttt{J}: No.}\\ \texttt{G}: Does the celebrity you're thinking of have a sister?  \user{\texttt{J}: Yes.}\\ \texttt{G}: Is the sister of your celebrity the same gender as the \\ \quad celebrity?  \user{\texttt{J}: Yes.}\\ \texttt{G}: Is the celebrity you're thinking of a musician?  \user{\texttt{J}: No.}\\ \texttt{G}: Is the celebrity you're thinking of a member of a pop  \\ \quad  group?  \user{\texttt{J}: No.}\\ \texttt{G}: The celebrity is not a musician, not a member of a pop  \\ \quad  group, and not a male. Do you know their name?  \user{\texttt{J}: Yes.}\\ \texttt{G}: You have provided the name of the celebrity. Is it a  \\ \quad  living person?  \user{\texttt{J}: Yes.}\\ \texttt{G}: Your celebrity is a living person who is not a musician,  \\ \quad  not a member of a pop group, and not a male. Can you  \\ \quad  provide their profession?  \user{\texttt{J}: Yes.}\\ ...}&
    \makecell[l]{\texttt{G}: Is the celebrity male?  \user{\texttt{J}: No.}\\ \texttt{G}: Is the celebrity a singer?  \user{\texttt{J}: No.}\\ \texttt{G}: Is the celebrity an actress?  \user{\texttt{J}: Yes.}\\ \texttt{G}: Has the celebrity won an Academy Award?  \user{\texttt{J}: Yes.}\\ \texttt{G}: Has the celebrity won the Academy Award for Best  \\ \quad  Actress?  \user{\texttt{J}: Yes.}\\ \texttt{G}: Has the celebrity won multiple Academy Awards for   \\ \quad  acting?  \user{\texttt{J}: No.}\\ \texttt{G}: Did the celebrity win the Academy Award for Best Actress  \\ \quad   for her role in a biographical film?  \user{\texttt{J}: Yes.}\\ \texttt{G}: Does the celebrity have blonde hair?  \user{\texttt{J}: No.}\\ \texttt{G}: Has the celebrity played a superhero character in a  \\ \quad   movie?  \user{\texttt{J}: Yes.}\\ \texttt{G}: Is the celebrity over the age of 40?  \user{\texttt{J}: No.}\\ \texttt{G}: Has the celebrity won Academy Awards for portraying  \\ \quad  real-life individuals on screen?  \user{\texttt{J}: Yes.}\\ \texttt{G}: Is the celebrity Charlize Theron?  \user{\texttt{J}: No.}\\ \texttt{G}: Is the celebrity Cate Blanchett?  \user{\texttt{J}: No.}\\ \texttt{G}: Is the celebrity Jennifer Lawrence?  \user{\texttt{J}: No.}\\ \texttt{G}: Is the celebrity Brie Larson?  \user{\texttt{J}: Bingo!}} \\
    \bottomrule
    \end{tabular}
    \caption{Comparison of gameplay scenario on \textit{Celebrities} between Vicuna 7B model vs \textbf{V-FT 7B (Things)}.}
    \label{tab:ft_vs_sc}
\end{table}

\section{GPT-4 failure cases}
\label{app:GPT_fail}
We notice that the strongest model GPT-4 consistently fail on the entity ``Yoga Mat'', while some other models can adeptly deduct it. We investigated this issue and it turns out that the GPT-4 model will always ask a misleading question ``Is it made of plastic'' in the 5 tested runs, leading to self-misinterpretation. See examples in Table~\ref{tab:gpt_fail}. GPT-4 seems to possess a stereotype of plastic items and excludes ``Yoga mat'' after knowing this item is made of plastic. The effect of this question is more evident in Table~\ref{tab:gpt_fail2}. After confirming that the entity is made of plastic, the model proceeds to focus on plastic only items in its top guesses for all following turns. This observation suggests that even the most robust model, which may demonstrate backtracking when encountering a wrong path, can still become trapped in erroneous assumptions.

\begin{table}[h]
    \centering
    \scriptsize
    \begin{tabular}{p{3.71in}|p{1.8in}}
    \toprule
    \multicolumn{2}{c}{Celebrity to be deducted: LeBron James} \\
    \hline
    \thead{Current top 4 guesses before generating question (Ranked)} & \thead{Generated question}\\
    \hline
    \begin{subtable}{0.6\linewidth}
    \rowcolors{0}{gray!25}{white} %
    \begin{tabular}{c|c|c|c|c}
    Leonardo DiCaprio & {Jennifer Aniston} & {Will Smith} & {Angelina Jolie} \\
    {Brad Pitt} & {Tom Cruise} & {Leonardo DiCaprio} & {Dwayne Johnson} \\
    {Justin Bieber} & {Drake} & Elon Musk & Cristiano Ronaldo \\
    {Donald Trump} & Elon Musk & Cristiano Ronaldo & {Richard Branson} \\
    {Michael Jordan} & {Tiger Woods} & \textcolor{blue}{LeBron James} & Cristiano Ronaldo \\
    Cristiano Ronaldo & \textcolor{blue}{LeBron James} & Lionel Messi & Tom Brady \\
    {Cristiano Ronaldo} & Lionel Messi & \textcolor{blue}{LeBron James} & {Tom Brady} \\
    \textcolor{blue}{LeBron James} & {Lionel Messi} & {Sidney Crosby} & Stephen Curry \\
    \textcolor{blue}{LeBron James} & Stephen Curry & Kevin Durant & {Kawhi Leonard} \\
    \textcolor{blue}{LeBron James} & {Stephen Curry} & {Kevin Durant} & {Giannis Antetokounmpo} \\
    \textcolor{blue}{LeBron James} & {Anthony Davis} & {Russell Westbrook} & {Carmelo Anthony} 

    \end{tabular}
    \end{subtable} &
    \makecell[l]{
        \texttt{G}: Is the celebrity male?  \user{\texttt{J}: Yes.}\\ 
        \texttt{G}: Is he an actor?  \user{\texttt{J}: No.}\\ 
        \texttt{G}: Is he a musician?  \user{\texttt{J}: No.}\\ 
        \texttt{G}: Is he involved in sports?  \user{\texttt{J}: Yes.}\\ 
        \texttt{G}: Is he a retired athlete?  \user{\texttt{J}: No.}\\ 
        \texttt{G}: Does he play a team sport?  \user{\texttt{J}: Yes.}\\ 
        \texttt{G}: Does he play American football?  \user{\texttt{J}: No.}\\ 
        \texttt{G}: Does he play basketball?  \user{\texttt{J}: Yes.}\\ 
        \texttt{G}: Is he currently playing in the NBA?  \user{\texttt{J}: Yes.}\\ 
        \texttt{G}: Does he play for the Los Angeles Lakers?  \user{\texttt{J}: Yes.}\\ 
        \texttt{G}: Is he LeBron James?  \user{\texttt{J}: Bingo!}
        }

    \\
    \bottomrule
    \end{tabular}
    \caption{
    Top 4 guesses made by GPT-4 during a game of guessing LeBron James. Guesses are generated before asking the clarification questions in the same line.
    }
    \label{tab:gpt_celeb_top_k}
\end{table}

\begin{table}[h]
    \centering
    \scriptsize
    \begin{tabular}{p{2.3in}|p{2.3in}}
    \toprule
    \thead{Repetition 1} & \thead{Repetition 2} \\
    \hline
    \makecell[l]{\texttt{G}: Is it a living entity?  \user{\texttt{J}: No.}\\ \texttt{G}: Is it man-made?  \user{\texttt{J}: Yes.}\\ \texttt{G}: Is it used for communication?  \user{\texttt{J}: No.}\\ \texttt{G}: Is it smaller than a bread box?  \user{\texttt{J}: Yes.}\\ \texttt{G}: Is it electronic?  \user{\texttt{J}: No.}\\ \texttt{G}: Is it made of metal?  \user{\texttt{J}: No.}\\ \texttt{G}: Is it used in daily life?  \user{\texttt{J}: Yes.}\\ \texttt{G}: Is it associated with personal hygiene?  \user{\texttt{J}: Maybe.}\\ \texttt{G}: Is it a container of sorts?  \user{\texttt{J}: No.}\\ \texttt{G}: Is it used in cooking or eating?  \user{\texttt{J}: No.}\\ \texttt{G}: Is it used for writing or drawing?  \user{\texttt{J}: No.}\\ \texttt{G}: Is it associated with clothing or fashion?  \user{\texttt{J}: No.}\\ \texttt{G}: Is it made of glass?  \user{\texttt{J}: No.}\\ \texttt{G}: \error{Is it made of plastic?}  \user{\texttt{J}: Yes.}\\ \texttt{G}: Is it found in a bathroom?  \user{\texttt{J}: No.}\\ \texttt{G}: Is it found in a bedroom?  \user{\texttt{J}: Maybe.}\\ \texttt{G}: Is it a toy?  \user{\texttt{J}: No.}\\ \texttt{G}: Is it used for storage?  \user{\texttt{J}: No.}\\ \texttt{G}: Is it used for cleaning purposes?  \\ \quad \user{\texttt{J}: No. You must guess now, what's it?}\\ \texttt{G}: Is it a plastic comb? Is it right?  \user{\texttt{J}: No.}} &
    \makecell[l]{\texttt{G}: Is it a living entity?  \user{\texttt{J}: No.}\\ \texttt{G}: Is it man-made?  \user{\texttt{J}: Yes.}\\ \texttt{G}: Is it something used daily?  \user{\texttt{J}: Maybe.}\\ \texttt{G}: Is it electronic?  \user{\texttt{J}: No.}\\ \texttt{G}: Is it something you can hold in your hand?  \user{\texttt{J}: Yes.}\\ \texttt{G}: Is it used for writing or drawing?  \user{\texttt{J}: No.}\\ \error{\texttt{G}: Is it made of plastic?}  \user{\texttt{J}: Yes.}\\ \texttt{G}: Is it used in the kitchen?  \user{\texttt{J}: No.}\\ \texttt{G}: Is it a toy?  \user{\texttt{J}: No.}\\ \texttt{G}: Is it used for personal care or hygiene?  \user{\texttt{J}: No.}\\ \texttt{G}: Is it used for storage?  \user{\texttt{J}: No.}\\ \texttt{G}: Is it a tool or instrument?  \user{\texttt{J}: Yes.}\\ \texttt{G}: Is it associated with a specific profession or hobby?  \\ \quad \user{\texttt{J}: Maybe.}\\ \texttt{G}: Is it used for measuring?  \user{\texttt{J}: No.}\\ \texttt{G}: Does it have moving parts?  \user{\texttt{J}: No.}\\ \texttt{G}: Is it used in crafting or DIY projects?  \user{\texttt{J}: Maybe.}\\ \texttt{G}: Is it a ruler?  \user{\texttt{J}: No.}\\ \texttt{G}: Is it a magnifying glass?  \user{\texttt{J}: No.}\\ \texttt{G}: Is it a stencil?  \user{\texttt{J}: No. You must guess now, what's it?}\\ \texttt{G}: Is it a glue stick? Is it right?  \user{\texttt{J}: No.}}
    \\
    \bottomrule
    \end{tabular}
    \caption{GPT-4 consistently fails on ``Yoga mat'' after confirming the entity is made of plastic.}
    \label{tab:gpt_fail}
\end{table}

\begin{table}[h]
    \centering
    \scriptsize
    \begin{tabular}{p{3.2in}|p{2.1in}}
    \toprule
    \multicolumn{2}{c}{Entity to be deducted: Yoga Mat} \\
    \hline
    \thead{Current top 3 guesses before generating question (Ranked)} & \thead{Generated question}\\
    \hline
    \begin{subtable}{0.6\linewidth}
    \rowcolors{0}{gray!25}{white} %
    \begin{tabular}{c|c|c|}
    {computer} & {tree} & {book} \\
    {Eiffel Tower} & {Great Pyramid of Giza} & {Statue of Liberty} \\
    book & {car} & {computer} \\
    {skyscraper} & book & smartphone \\
    {smartphone} & {car} & book \\
    book & {chair} & {table} \\
    {pencil} & book & {pair of glasses} \\
    {pen} & {book} & {wine glass}  \\
    {wooden spoon} & {glass cup} & plastic comb  \\
    {plastic water bottle} & {plastic comb} & plastic spoon  \\
    {plastic spoon} & {plastic cup} & plastic toy \\
    plastic pen & plastic ruler & {plastic toy} \\
    {plastic pen} & plastic comb & {plastic ruler} \\
    plastic comb & plastic spoon & plastic toy car \\
    {plastic comb} & {plastic spoon} & {plastic toy car} \\
    plastic storage box & plastic organizer drawer & {plastic file folder}  \\
    plastic box for storage & {plastic hanger} & {plastic drawer organizer}  \\
    plastic box or container & plastic toy bin & {plastic shoe rack}  \\
    {plastic storage box} & {plastic toy box} & {plastic jewelry box} \\
    plastic cup & plastic ruler & plastic toy 
    \end{tabular}
    \end{subtable} &
    \makecell[l]{
        \texttt{G}: Is it a living entity?  \user{\texttt{J}: No.}\\ 
        \texttt{G}: Is it a living tangible?  \user{\texttt{J}: Yes.}\\ 
        \texttt{G}: Is it a man-made?  \user{\texttt{J}: Yes.}\\ 
        \texttt{G}: Can it be used in everyday life?  \user{\texttt{J}: Yes.}\\ 
        \texttt{G}: Is it electronic?  \user{\texttt{J}: No.}\\ 
        \texttt{G}: Is it smaller than a bread box?  \user{\texttt{J}: Yes.}\\ 
        \texttt{G}: Is it made of metal?  \user{\texttt{J}: No.}\\ 
        \texttt{G}: Is it made out of paper?  \user{\texttt{J}: No.}\\ 
        \error{\texttt{G}: Is it made of plastic?}  \user{\texttt{J}: Yes.}\\ 
        \texttt{G}: Is it used for personal hygiene?  \user{\texttt{J}: No.}\\ 
        \texttt{G}: Is it used in the kitchen?  \user{\texttt{J}: No.}\\ 
        \texttt{G}: Is it used for entertainment or leisure?  \user{\texttt{J}: No.}\\ 
        \texttt{G}: Is it used in an office setting?  \user{\texttt{J}: No.}\\ 
        \texttt{G}: Is it related to education or learning?  \user{\texttt{J}: No.}\\ 
        \texttt{G}: Is it associated with storage or organization?  \user{\texttt{J}: Maybe.}\\ 
        \texttt{G}: Is it found in a bedroom?  \user{\texttt{J}: Maybe.}\\ 
        \texttt{G}: Is it associated with clothing or fashion?  \user{\texttt{J}: No.}\\ 
        \texttt{G}: Is it typically found in a garage or outdoors?  \user{\texttt{J}: No.}\\ 
        \texttt{G}: Is it used to hold or contain smaller items?  \user{\texttt{J}: No.}\\ 
        \texttt{G}: Is it a plastic ruler?  \user{\texttt{J}: No.}
}

    \\
    \bottomrule
    \end{tabular}
    \caption{
    Top 3 guesses made by GPT-4 guessing entity ``Yoga Mat''. Following the question in red, all top guesses focus on plastic.
    }
    \label{tab:gpt_fail2}
\end{table}

\end{document}